# Greedier is Better: Selecting Multiple Neighbors per Iteration for Sparse Subspace Clustering

Jwo-Yuh Wu*, Liang-Chi Huang, Wen-Hsuan Li, Chun-Hung Liu, and Rung-Hung Gau

***Abstract*—Sparse subspace clustering (SSC) using greedy-based neighbor selection, such as orthogonal matching pursuit (OMP), has been known as a popular computationally-efficient alternative to the popular $\ell_1$-minimization based methods. This paper proposes a new SSC scheme using generalized OMP (GOMP), a soup-up of OMP whereby multiple neighbors are identified per iteration, along with a new stopping rule requiring nothing more than a knowledge of the ambient signal dimension. Compared to conventional OMP, which identifies one neighbor per iteration, the proposed GOMP method involves fewer iterations, thereby enjoying lower algorithmic complexity; advantageously, the proposed stopping rule is free from off-line estimation of subspace dimension and noise power. Under the semi-random model, analytic performance guarantees, in terms of neighbor recovery rates, are established to justify the advantage of the proposed GOMP. The results show that, with a high probability, GOMP (i) is halted by the proposed stopping rule, and (ii) can retrieve more true neighbors than OMP, consequently yielding higher final data clustering accuracy. Computer simulations using both synthetic data and real human face data are provided to validate our analytic study and evidence the effectiveness of the proposed approach.**

## I. Introduction

### A. Motivation

Subspace clustering is a key enabling technique in modern unsupervised machine learning [1-3]. Formally, consider a noisy data set $\mathcal{Y}=\{\mathbf{y}_1,\cdots,\mathbf{y}_N\}\subset\mathbb{R}^n$ whose ground truth obeys a disjoint union as

$$\mathcal{Y}=\mathcal{Y}_1\cup\mathcal{Y}_2\cup\cdots\cup\mathcal{Y}_L, \tag{1.1}$$

where cluster $\mathcal{Y}_k\subset\mathbb{R}^n$ consists of $|\mathcal{Y}_k|>0$ noisy data points coming from a $d_k$-dimensional subspace $\mathcal{S}_k$, and $|\mathcal{Y}_1|+\cdots+|\mathcal{Y}_L|=N$. A partition of $\mathcal{Y}$ into the form (1.1) is widely known as the union-of-subspaces model [1], which underpins a panoply of practical data clusters ranging from human face images, hand written digits, to trajectories of moving objects in videos. Given only the data set $\mathcal{Y}$ and without knowing the size $L$ of partition, subspaces $\mathcal{S}_k$'s and their dimensions $d_k$'s, and noise level, the task of subspace clustering is to unearth the partition (1.1). Among existing solutions to this problem, sparse subspace clustering (SSC) [4-7], catalyzed by the sheer success of compressive sensing (CS) and

This work is sponsored by the Ministry of Science and Technology of Taiwan under grants MOST 108-2221-E-009-025 MY3 and MOST 108-2634-F-009-002.

J. Y. Wu, L. C. Huang, W. H. Li, and and R. H. Gau are with the Institute of Communications Engineering, the Department of Electrical and Computer Engineering, and College of Electrical Engineering, National Yang Ming Chiao Tung University, 1001, Ta-Hsueh Road, Hsinchu, Taiwan. Emails: jywu@cc.nctu.edu.tw; uuranus001@gmail.com; fiend_danger@hotmail.com; runghung-gau@g2.nctu.edu.tw.

C. H. Liu is with the Department of Electrical and Computer Engineering, Mississippi State University, USA. E-mail: chliu@ece.msstate.edu.

* Contact author.

sparse representation [8-11], has gained much attention of late, not least because of its compelling experimental performance and provable performance guarantees [12-20]. Typically, SSC first resorts to sparse regression for neighbor identification, then to construct a similarity graph of the given data set, followed by spectral clustering for final data segmentation. Perhaps the most popular sparse regression for SSC is the $\ell_1$ -minimization approach and its variants thereof [7], which are however computationally demanding in practical applications. Accordingly, low-complexity alternatives using greedy-based neighbor selection, e.g., orthogonal matching pursuit (OMP) [10, 11], have been proposed, yet able to perform on par with $\ell_1$ -minimization in many cases.

Added to the reams of algorithm development for SSC are vigorous studies of mathematical performance guarantees, thanks to the fruitful analytical tools in CS. The vast majority of related works centered around investigating sufficient conditions ensuring the so-called subspace detection property (SDP) [12], i.e., sparse regression certainly returns a neighbor subset from the correct cluster, therefore producing a similarity graph without inter-cluster edges; see [12-16] for discussions regarding the $\ell_1$ -minimization solutions, and [17-20] pertinent to greedy search. From the perspective of network connectivity, errorless neighbor identification alone as claimed by SDP is by and large an overly pessimistic guideline, on the ground that SDP is not necessary for perfect data segmentation. Indeed, as reported in many studies [21-22], a known type of similarity graphs effectuating successful clustering is one configured with many intra-cluster and few inter-cluster edges. Evidently, this arises when the sparse regression mis-identifies few neighbors, violating SDP. The downside of few falsely directed edges from cluster to cluster can be effectively compensated by spectral clustering (see [23] for more discussions on this issue). Be that as it may, occasions when matters grow worse would come about, especially that SDP is accompanied by meager neighbors. The proximate cause is, despite no inter-cluster edges, the resultant similarity graph is cut into excessively many isolated pieces, poor graph connectivity in this way begetting an over-estimation of the number $L$ of clusters, leading to the failure of data clustering [13, 14]. The above facts altogether shad further light on the study of sparse regression for SSC. On the algorithm design aspect, the efforts shall be particularly geared towards fast acquisition of plentiful neighbors, hopefully many true, so that the similarity graph can be quickly fleshed out in a right configuration. As to neighbor recovery performance guarantees, much remains to be explored in quest to new leitmotivs not so stringent as SDP, especially one able to reflect neighbor recovery error. Under the framework of two-step weighted $\ell_1$ -minimization, [14] analyzed the neighbor recovery rate, specifically, the probability that the sparse regressor produces at least $k_t(>0)$ correct and at most $k_f(\geq 0)$ incorrect neighbors. A probabilistic characterization of this sort is intuitive and quite flexible, directly taking account of the general case with neighbor mis-identification; when specializing to $k_f = 0$, i.e., errorless neighbor identification, it then reveals how much chance SDP stands, with no less than $k_t$ recovered neighbors.

### *B. Paper Contributions*

This paper aims at tackling the aforesaid challenge by revamping the OMP, considering its up-to-par performance, reduced computational complexity and, most importantly, the inherent flexibility to boost recovered neighbors. Pivoted on its prominent variant, the generalized orthogonal matching pursuit (GOMP), first introduced in the literature of CS [24] and allowed to identify multiple neighbors per iteration, we propose a new sparse regression scheme for SSC, together with an in-depth analytic study of the neighbor recovery performance. Specific technical contributions of this paper can be summarized as follows.

1. We propose to employ GOMP as an effective alternative to OMP for fast acquiring plentiful neighbors, with the prospect of many true. Particularly, we first point out the deviation (caused by noise) of the residual vector from the desired ground truth subspace, pinned down by the Angle of Deviation (AoD), plays a pivotal role in neighbor identification. Helped by this, we then argue that GOMP, while digging out more neighbors per iteration, enjoys a smaller AoD, thus better able to withstand noise corruption and achieve potentially higher neighbor identification accuracy than OMP.
2. Efficient stopping rules are crucial for greedy neighbor search. If the algorithm stops too early, we would end up with scant correct neighbors, yet if too late, with overly many false ones; either case is apt to cause poor graph connectivity and eventual erroneous data clustering. Alongside the proposed GOMP we devise a new stopping rule to halt neighbor selection. By examining the way residual power dissipates from iteration to iteration when left wholly with noise, a right moment to leave off, our stopping rule is geared to be self-aware of this situation. Mathematically, it amounts to judging the ratio of the residual norms in consecutive two iterations against a threshold depending solely on the ambient space dimension $n$, which is known once the data set is given. Advantageously, this dispenses with extra off-line estimation of the subspace dimension or background noise power, as required in the existing solution [13-20].
3. Capitalized on our previous work [14] and under the semi-random model [12-13], we conduct recovery rate analysis to bear out the claimed merits of the proposed GOMP scheme. Supposing that GOMP recovers $p(\geq 1)$ neighbors in each iteration, we first derive an analytic probability lower bound for the event that at least $k_m$ $(0 \leq k_m \leq p)$ are correct in the $m$th iteration. Such an "iteration-wise" recovery rate result is then exploited to obtain the global recovery rate, the one corresponding to the situation that at least $k_t$ correct neighbors in total are identified once the search ends. The obtained analytic formula confirms that, for a large data size $N$ and small noise power, GOMP enjoys a higher correct neighbor recovery rate than OMP, thereby supporting GOMP betters fast identification of many correct neighbors. Finally, we show that, with a high probability, the GOMP algorithm is terminated according to the proposed stopping rule.

The rest of this paper is organized as follows. Section 2 first explains why GOMP can outdo conventional OMP, and introduces the foundations behind the proposed stopping criterion. Section III presents the recovery rate analyses. Section IV provides numerical simulations to verify our theoretical study and discussions in Section III. Section V goes through the proofs of the main mathematical results. Finally, Section VI concludes this paper. To ease reading, some detailed proofs are relegated to the appendix.

# II. Proposed SSC-GOMP

This section introduces the proposed scheme. We first brief in Section II-A the reason why multiple neighbors per iteration is favored, in an attempt to motivate our GOMP proposal. In Section II-B, we then encapsulate the foundations behind the proposed stopping rule, then ending with an outline of the proposed SSC-GOMP algorithm.

## *A. Why Multiple Neighbors per Iteration?*

Recall that OMP iteratively identifies a neighbor each time as the data point when paired with the residual vector yields peak absolute inner product. Hence, the orientation of the residual vector in each iteration, in particular, the degree to which it deviates from the ground truth subspace, is a whole lot important for accurate neighbor identification. To formalize this notion, assume that we are to build a neighbor list for the data point $\mathbf{y}_i$ lying in subspace $\mathcal{S}_k$. Impaired by noise, the residual vector $\mathbf{r}_m^{(i)}$ computed in the $m$th iteration ($m \geq 1$) is perturbed outwards $\mathcal{S}_k$. If we write $\mathbf{r}_m^{(i)} = \mathbf{r}_{m,\|}^{(i)} + \mathbf{r}_{m,\perp}^{(i)}$, where $\mathbf{r}_{m,\|}^{(i)} \in \mathcal{S}_k$ and $\mathbf{r}_{m,\perp}^{(i)} \in \mathcal{S}_k^{\perp}$ (the orthogonal complement of $\mathcal{S}_k$), such perturbation can be pinned down by the *angle of deviation* (AoD)

$$\phi_m^{(i)} \triangleq \tan^{-1}\left(\frac{\left\|\mathbf{r}_{m,\perp}^{(i)}\right\|_2}{\left\|\mathbf{r}_{m,\|}^{(i)}\right\|_2}\right), \tag{2.1}$$

whereby a large $\phi_m^{(i)}$ means $\mathbf{r}_m^{(i)}$ severely deviates from $\mathcal{S}_k$. As the OMP algorithm crunches with more iterations, the residual has a propensity to point more and more astray from $\mathcal{S}_k$. This is mainly because, owing to orthogonal projection (see step 3 in Table I), the current residual is obtained from the previous one by removing from it the component lying in the subspace spanned by the already selected neighbors, most being correct with a high chance. Consequently, the magnitude $\left\|\mathbf{r}_{m,\|}^{(i)}\right\|_2$ of the component $\mathbf{r}_{m,\|}^{(i)} \in \mathcal{S}_k$ diminishes from iteration to iteration; instead, the term $\left\|\mathbf{r}_{m,\perp}^{(i)}\right\|_2$, which reflects the accumulative noise strength, is typically non-decreasing with $m$. Put together, the cascade effect is therefore an increase in $\phi_m^{(i)}$ with $m$, rendering the residual $\mathbf{r}_m^{(i)}$ iteration after iteration increasingly prone to neighbor mis-identification. In this regard, a simple remedy for securing enough correct neighbors in few iterations (a "small $m$" is favored) is therefore to dig out multiple neighbors per iteration, say, the $p$ data points ($p > 1$) corresponding to the largest $p$ absolute inner products: GOMP is therefore in with a chance. Using synthetic data, Fig. 1-(a) clearly demonstrates GOMP yields smaller AoD than OMP thanks to fewer iterations; this accordingly brings about higher neighbor identification accuracy, as illustrated in Fig 1-(b) (this issue will be elaborated more in Section III).

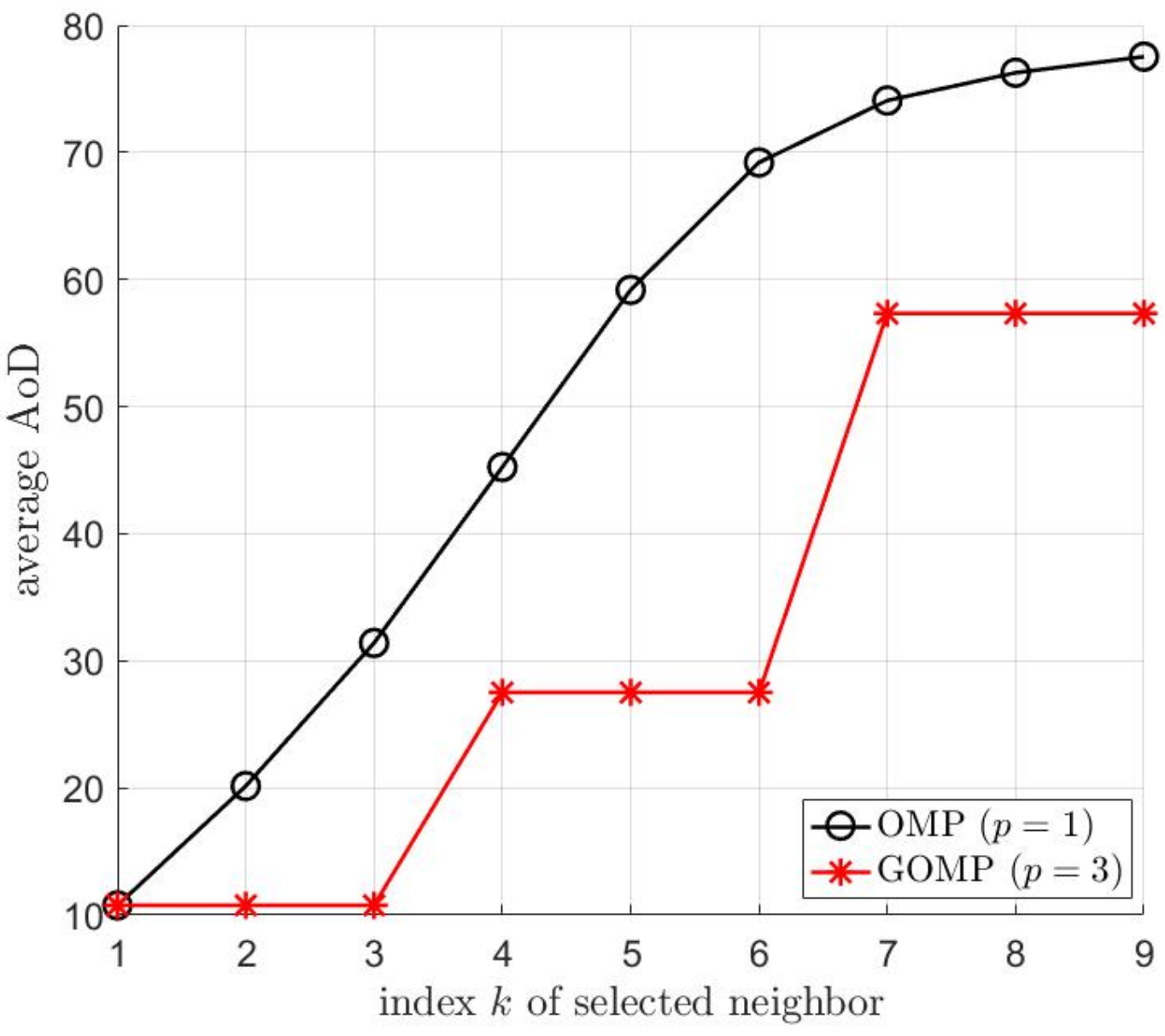

Fig. 1-(a). Average AoD versus neighbor index $k$ with $p = 1$ and $p = 3$.

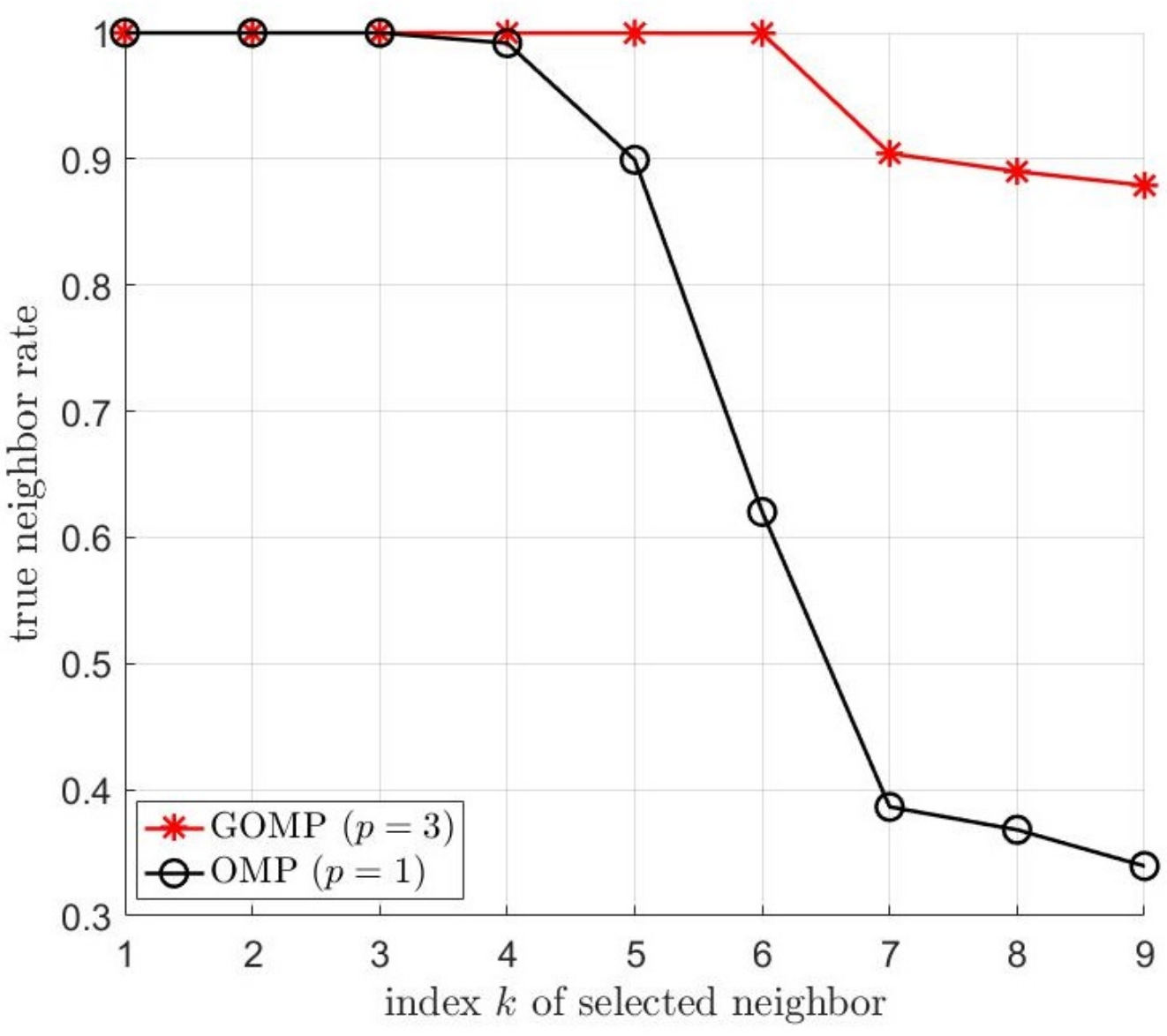

Fig. 1-(b). true neighbor rate versus neighbor index $k$ with $p = 1$ and $p = 3$.

Fig. 1. Comparison of GOMP and OMP in terms of average AoD and empirical recovery rate. Consider a synthetic data set of $135$ vectors drawn from $L = 3$ orthogonal subspaces, each of a dimension $9$, in an ambient domain $\mathbb{R}^{100}$; $45$ data points per cluster. The data vectors are sampled uniformly from the intersection of the unit-sphere in $\mathbb{R}^{100}$ with the ground truth subspace, and are corrupted by zero mean Gaussian noise with variance $0.04$. For GOMP, $p = 3$ neighbors are picked per iteration as those when matched to the residual yielding the largest three absolute inner products. A total number of $9$ neighbors are recovered using both OMP and GOMP. (a) Plot of average AoD upon detection of the $k$th neighbor, $1 \leq k \leq 9$. For GOMP, every three neighbors are detected in each iteration based on the same residual, leading to a staircase AoD curve. Clearly, GOMP results in smaller AoD thanks to fewer iterations. (b) Plot of the true neighbor rate, i.e., the fraction of true neighbors recovered, versus the index $k$ of the detected neighbor. Benefiting from smaller AoD, GOMP is seen to improve neighbor identification accuracy: even for the last neighbor, it allows a true neighbor rate about as high as $0.9$, up from less than $0.4$ of OMP.

Another upside from fewer iterations is lower algorithmic complexity, which is potentially appealing in real-time applications. The major computational bottleneck of the OMP algorithm is the orthogonal projection operation (step 3 in Table I, with $p=1$) that distils the innovation out from the already selected neighbors. To recover $p$ neighbors, conventional OMP requires $p$ iterations, hence $p$ orthogonal projections. Instead, GOMP calls for just one iteration, so one orthogonal projection only, and is therefore more computationally efficient. Considering all the above facts, we therefore propose to adopt GOMP in place of OMP for neighbor identification.

*B. Halt When There Are About as Many Neighbors as Subspace Dimension*

Since the data point $\mathbf{y}_i$ belongs to $\mathcal{S}_k$, which is $d_k$-dimensional, a group of around $d_k$ true neighbors oftentimes reaches a "critical mass" to well explain $\mathbf{y}_i$ and, if so, the residual from now on is highly apt to be dominated by noise, standing very little chance to uncover more true neighbors (this will be born out by our mathematical analysis in Section V). Grounded on this fact, the algorithm is expected to halt once $d_k$ neighbors or so are available. We shall recall the stopping rule widely adopted in the literature [19, 20], which terminates neighbor search when either the number of iterations reaches a pre-set bound $M$, or the residual becomes so small that $\left\|\mathbf{r}_m^{(i)}\right\|_2 \le \tau$ for some threshold $\tau > 0$. Though implicit, this imposes additional off-line initialization tasks of determining the parameters $M$ and $\tau$ (or side information thereof) that, however, could be costly or even impossible. Indeed, $M$ is arguably all about the subspace dimension $d_k$, which is nonetheless unknown; $\tau$ is closely related to the background noise level yet to be acquired through extra training and measurement process. Heedful of the above concerns, below we develop a new stopping rule which is *per se* aware of the subspace dimension and, in turn, eradicates the initialization tasks as required by the solution in [19, 20].

The proposed approach is premised on the following lines of thinking. Suppose that the residual $\mathbf{r}_m^{(i)}$ in the $m$th iteration is depleted so that it consists of noise only; this typically happens once there are about as many recovered neighbors as the subspace dimension $d_k$. When the background noise corruption is Gaussian, so is the noise-only residual $\mathbf{r}_m^{(i)}$, which, being nearly isotropic, tends to distribute its power evenly over all the ambient dimensions, about $\left\|\mathbf{r}_m^{(i)}\right\|_2 / \sqrt{n}$ each when taking square root. During the $(m+1)$th iteration, $\mathbf{r}_{m+1}^{(i)}$ is then obtained from $\mathbf{r}_m^{(i)}$ by removing from it the components along the $p$ newly selected neighbors, implying that $\left\|\mathbf{r}_m^{(i)} - \mathbf{r}_{m+1}^{(i)}\right\|_2$ is close to $\sqrt{p} \times \left\|\mathbf{r}_m^{(i)}\right\|_2 / \sqrt{n}$. Using triangle inequality, we then come up with the following condition to halt neighbor search

$$\left\|\mathbf{r}_m^{(i)}\right\|_2 - \left\|\mathbf{r}_{m+1}^{(i)}\right\|_2 \le \left(\left\|\mathbf{r}_m^{(i)}\right\|_2 \sqrt{p}\right) / \sqrt{n}\,, \tag{2.2}$$

or equivalently,

$$\frac{\left\|\mathbf{r}_{m+1}^{(i)}\right\|_2}{\left\|\mathbf{r}_m^{(i)}\right\|_2} \ge 1 - \sqrt{p/n}\,. \tag{2.3}$$

Table I. Outline of the proposed SSC-GOMP algorithm.

| SSC-GOMP algorithm with the proposed data dependent stopping criterion |
|---|
| **Input:** Observed data set $\mathcal{Y}=\{\mathbf{y}_1,\cdots,\mathbf{y}_N\}$. |
| **For** $i=1,\cdots N$<br>Let $m=0$, $\mathbf{r}_0=\mathbf{y}_i$, $\mathbf{r}_{-1}=2\mathbf{y}_i$, and $\Lambda_0=\phi$.<br>**Do if** ( $\left(1-\lVert\mathbf{r}_m\rVert/\lVert\mathbf{r}_{m-1}\rVert\right)\geq\sqrt{p/n}$ )<br>1. $m=m+1$.<br>2. $\Lambda_m=\Lambda_{m-1}\cup\mathcal{T}_m$, where $\mathcal{T}_m$ is the set of cardinality $p$ such that<br>$\left\lvert\left\langle\mathbf{y}_j,\mathbf{r}_{m-1}\right\rangle\right\rvert\geq\left\lvert\left\langle\mathbf{y}_q,\mathbf{r}_{m-1}\right\rangle\right\rvert,\ \forall j\in\mathcal{T}_m,\ q\notin\mathcal{T}_m$.<br>3. $\mathbf{r}_m=(\mathbf{I}-\mathbf{Y}_{\Lambda_m}\mathbf{Y}_{\Lambda_m}^{\dagger})\mathbf{r}_{m-1}$.<br>**end**<br>4. $\mathbf{c}_i^*=\underset{\mathbf{c}:\mathrm{supp}(\mathbf{c})\subset\Lambda_{m\text{-}1}}{\arg\min}\lVert\mathbf{y}-\mathbf{Y}\mathbf{c}\rVert_2$.<br>5. Normalize the column vector $\mathbf{c}_i^*$ to be unit-norm, and let<br>$\bar{\mathbf{c}}_i^*=[c_{i,1}^*\cdots c_{i,i-1}^*\ 0\ c_{i,i+1}^*\cdots c_{i,N}^*]^T\in\mathbb{R}^N$.<br>**end**<br>6. Set $\mathbf{C}=[c_{i,j}]=[\bar{\mathbf{c}}_1^*\cdots\bar{\mathbf{c}}_N^*]$, and $\mathbf{G}=[g_{i,j}]$, where $g_{i,j}=\lvert c_{i,j}\rvert+\lvert c_{j,i}\rvert$.<br>7. Form a similarity graph with $N$ nodes, with the weight on the edge between the ($i$, $j$) node pair equal to $g_{i,j}$.<br>8. Apply spectral clustering to the similarity graph. |
| **Output:** Partition $\mathcal{Y}=\widehat{\mathcal{Y}}_1\cup\cdots\cup\widehat{\mathcal{Y}}_{\hat{L}}$ |

The proposed halting rule is dimensionality-aware in the sense that, for most cases, only the depletion of $\mathbf{r}_m^{(i)}$ thanks to a pile of $d_k$ recovered neighbors or thereabouts can trigger the condition (2.3). As against the criterion in [19, 20], our solution (2.3) compares the ratio of consecutive two residual norms against a threshold assuming nothing more than a knowledge of the ambient dimension $n$, which is always known in advance, and the parameter $p$ entirely of the designer's disposal; it is therefore untethered from extra off-line burdens of subspace dimension and noise power estimation. We summarize the proposed SSC-GOMP algorithm in Table I.

# III. THEORETICAL RESULTS

In this section, we present the recovery rate analysis for the proposed SSC-GOMP. To formalize matters, the data vector is assumed to follow the standard additive noise model, that is,

$$\mathbf{y}_i=\mathbf{x}_i+\mathbf{e}_i,\ \ 1\leq i\leq N, \tag{3.1}$$

where $\mathbf{x}_i\in\mathbb{R}^n$ is the unit-norm noiseless signal vector and $\mathbf{e}_i\in\mathbb{R}^n$ is the noise. The analyses below are built on the popular semi-random model [12-16], i.e., the ground truth subspaces $\mathcal{S}_1,\cdots,\mathcal{S}_L$ in the partition (1.1) are fixed but otherwise unknown, whereas the data vectors and noise are random. Such a model is widely used in the theoretical study of SSC, thanks to its interpretability and amiability to analysis. Similar to the previous works [13, 14, 18, 20], the following assumptions are made in the sequel.

***Assumption 1:*** For each $1 \le i \le N$, the signal vector $\mathbf{x}_i \in \mathcal{S}_k$ is uniformly sampled from $\mathbf{x}_i \in \mathcal{B}_k$, where $\mathcal{B}_k \triangleq \{\mathbf{x} \mid \mathbf{x} \in \mathbb{R}^n, \|\mathbf{x}\|_2 = 1\} \cap \mathcal{S}_k$ is the intersection of the unit sphere with the subspace $\mathcal{S}_k$. □

***Assumption 2:*** For each $1 \le i \le N$, the noise $\mathbf{e}_i \in \mathbb{R}^n$ are i.i.d. Gaussian random vectors with zero mean and covariance matrix $(\sigma^2 / n)\mathbf{I}$, i.e., $\mathbf{e}_i \sim \mathcal{N}(0, \sigma^2 / n\mathbf{I})$, and are independent of the signal vectors $\mathbf{x}_i$'s. □

To gauge the degree to which two subspaces are separated away from each other, we recall the affinity between two distinct subspaces $\mathcal{S}_k$ and $\mathcal{S}_l$ that is defined to be [12]

$$aff\,(\mathcal{S}_k, \mathcal{S}_l) \triangleq \frac{\|\mathbf{U}_k^T \mathbf{U}_l\|_F}{\sqrt{\min\{d_k, d_l\}}}, \tag{3.2}$$

where columns of $\mathbf{U}_k$ ($\mathbf{U}_l$, respectively) form an orthonormal basis for $\mathcal{S}_k$ ($\mathcal{S}_l$, respectively).

***Assumption 3:*** The subspace affinity $aff\,(\mathcal{S}_k, \mathcal{S}_l)$ satisfies

$$\max_{k:k \ne l} aff\,(\mathcal{S}_k, \mathcal{S}_l) + \frac{9\sqrt{3}d_L}{(8 - 12\sigma)\sqrt{(n - d_L)\log N}}(1 + \sigma) \le \frac{\tau}{4\log N}, \tag{3.3}$$

where $0 < \tau < 1$. □

Notably, Assumptions 3 guarantees that different subspaces are well separated from each other; affinity conditions akin to (3.3) are also needed in many existing studies of performance guarantees for SSC [13, 14, 18, 20]. Without loss of generality, we assume in the sequel $\mathbf{x}_N \in \mathcal{S}_L$, therefore $\mathbf{y}_N \in \mathcal{Y}_L$, and our goal is to identify a neighbor group of $\mathbf{y}_N$. Under the above three assumptions it can be shown that, with a large data size and small noise corruption, GOMP stands a high chance to recover many true neighbors per iteration up to $\lceil d_L / p \rceil$ rounds, whence GOMP is about to halt as argued in Section II-B. More precisely, we have the following theorem.

***Theorem 3.1 (Iteration-Wise Recovery Rate):*** Let $\{k_1, k_2, \cdots k_M\}$ be a sequence of integers satisfying $0 \le k_m \le p$, for each $1 \le m \le M \le \lceil d_L / p \rceil$. Under Assumptions 1~3, the proposed SSC-GOMP obtains at least $k_m$ true neighbors at the $m$th iteration, $1 \le m \le M$, with a probability exceeding

$$\begin{aligned} &1 - Ne^{-n/8} - v(d_L - p(M-1))\left(\frac{\sigma}{\sqrt{\pi}}\right)^{d_L - p(M-1)} \\ &-\sum_{m=1}^{M}\left[\left(\left(\frac{2e(N - |\mathcal{Y}_L|)}{(p - k_m + 1)N^{8\log N / d_L}}\right)^{(p-k_m+1)} + \left(\sqrt{\frac{2}{\pi}}\tau\right)^{(|\mathcal{Y}_L| - d_L - k_m)}\left(\frac{e(|\mathcal{Y}_L| - 1)}{k_m - 1}\right)^{k_m - 1} + \frac{4 + 2c}{N^2}\right)\mathbf{1}(k_m > 0)\right], \end{aligned} \tag{3.4}$$

where $\mathbf{1}(\cdot)$ is the indicator function, $c$ is a positive constant, $d_L$ is the dimension of the subspace $\mathcal{S}_L$, and $v(d_L)$ is the volume of unit-ball in $\mathbb{R}^{d_L}$.

*[Proof]:* See Section V-A. □

Theorem 3.1 provides an analytic probability lower bound for the event that GOMP succeeds in identifying at least $k_m$ true neighbors in the $m$th iteration. A further scrutiny reveals the lower bound (3.4) is high in most practical cases. Indeed, with a large ambient dimension $n$ and small noise level $\sigma$, the second and third terms in (3.4) are kept small; regarding the last summation, the first and third terms scale like $N^{[1-(8\log N/d_L)](p-k_m+1)}$ and $N^{-2}$, respectively, whereas the second term decays exponentially fast as $|\mathcal{Y}_L|$ increases, thanks to $0<\tau<1$ (see Assumption 3). When specialized to $p=1$ and $k_m=1$ for all $1\le m\le M$, i.e., the case with conventional OMP subject to all detected neighbors being true, the lower bound (3.4) then reads

$$1-Ne^{-n/8}-v(d_L-(M-1))\left(\frac{\sigma}{\sqrt{\pi}}\right)^{d_L-(M-1)}-M\left[\frac{2e(N-|\mathcal{Y}_L|)}{N^{8\log N/d_L}}+\left(\sqrt{\frac{2}{\pi}}\tau\right)^{(|\mathcal{Y}_L|-d_L-1)}+\frac{4+2c}{N^2}\right]. \tag{3.5}$$

Notably, a probability lower bound akin to (3.5) was also reported in [20], which addressed sufficient conditions ensuring correct neighbor recovery. Based on the iteration-wise recovery rate result of Theorem 3.1, the following corollary further establishes the global recovery rate, namely, the probability of the event that GOMP, when leaving off, succeeds in recovering a specified total number of true neighbors.

***Corollary 3.2 (Global Recovery Rate):*** Let $0\le k_t\le pM$ be an integer and write $k_t=Mq_t+r_t$, where $0\le r_t\le M-1$. Under the same setup as in Theorem 3.1, GOMP can recover at least totally $k_t$ true neighbors throughout $M(\le\lfloor d_L/p\rfloor)$ iterations with a probability higher than

$$\begin{aligned}
&1-Ne^{-n/8}-v(d_L-p(M-1))\left(\frac{\sigma}{\sqrt{\pi}}\right)^{d_L-p(M-1)}\\
&-r_t\left[\left(\frac{2e(N-|\mathcal{Y}_L|)}{(p-q_t)N^{8\log N/d_L}}\right)^{(p-q_t)}+\left(\sqrt{\frac{2}{\pi}}\tau\right)^{(|\mathcal{Y}_L|-d_L-q_t-1)}\left(\frac{e(|\mathcal{Y}_L|-1)}{q_t}\right)^{q_t}+\frac{4+2c}{N^2}\right]\\
&-(M-r_t)\left[\left(\left(\frac{2e(N-|\mathcal{Y}_L|)}{(p-q_t+1)N^{8\log N/d_L}}\right)^{(p-q_t+1)}+\left(\sqrt{\frac{2}{\pi}}\tau\right)^{(|\mathcal{Y}_L|-d_L-q_t)}\left(\frac{e(|\mathcal{Y}_L|-1)}{q_t-1}\right)^{q_t-1}+\frac{4+2c}{N^2}\right)\mathbf{1}(q_t>0)\right].
\end{aligned} \tag{3.6}$$

*[Proof]:* See Section V-B. □

By following our examination of (3.4), it is easy to check the probability lower bound (3.6) is high in most practical cases. The bound (3.6) can moreover buttress the advantage of multiple neighbor recovery of GOMP as compared with the conventional OMP. To better illustrate this, we assume without loss of generality that a group of $pM$ neighbors is to be found, while demanding at least $k_t=kM$ true neighbors, for some $1<k\le p$ (that is to say, $k/p$ of the total neighbors are true). Hence, GOMP requires $M$ iterations, and OMP $p$ times more. Under this setting, the lower bound (3.6) for GOMP reads

$$1-Ne^{-n/8}-v(d_L-pM+p)\left(\frac{\sigma}{\sqrt{\pi}}\right)^{d_L-pM+p}-M\left(\frac{2e(N-|\mathcal{Y}_L|)}{(p-k+1)N^{8\log N/d_L}}\right)^{(p-k+1)}$$
$$-M\left(\sqrt{\frac{2}{\pi}}\tau\right)^{(|\mathcal{Y}_L|-d_L-k)}\left(\frac{e(|\mathcal{Y}_L|-1)}{k-1}\right)^{k-1}-\frac{(4+2c_2)M}{N^2}, \tag{3.7}$$

whereas for OMP it becomes

$$1-Ne^{-n/8}-v\left(d_L-pM+1\right)\left(\frac{\sigma}{\sqrt{\pi}}\right)^{d_L-pM+1}$$
$$-\frac{2kMe(N-|\mathcal{Y}_L|)}{N^{8\log N/d_L}}-kM\left(\sqrt{\frac{2}{\pi}}\tau\right)^{(|\mathcal{Y}_L|-d_L-1)}-\frac{(4+2c)kM}{N^2}. \tag{3.8}$$

Note that (3.7) differs from (3.8) in the last four terms, among which it is clear that $v(d_L-pM+p)\left(\frac{\sigma}{\sqrt{\pi}}\right)^{d_L-pM+p}<v(d_L-pM+1)\left(\frac{\sigma}{\sqrt{\pi}}\right)^{d_L-pM+1}$ for small noise level $\sigma$, and $\frac{(4+2c)M}{N^2}<\frac{(4+2c)kM}{N^2}$ because $k>1$. For a sufficiently large data size $N$, the terms $M\left(\frac{2e(N-|\mathcal{Y}_L|)}{(p-k+1)N^{8\log N/d_L}}\right)^{(p-k+1)}$ and $\frac{2kMe(N-|\mathcal{Y}_L|)}{N^{8\log N/d_L}}$ scale like $N^{[1-(8\log N/d_L)](p-k+1)}$ and $N^{[1-(8\log N/d_L)]}$, respectively, and thus both vanish. Also, since $\tau<1$, both $M\left(\sqrt{\frac{2}{\pi}}\tau\right)^{(|\mathcal{Y}_L|-d_L-k)}\left(\frac{e(|\mathcal{Y}_L|-1)}{k-1}\right)^{k-1}$ and $kM\left(\sqrt{\frac{2}{\pi}}\tau\right)^{(|\mathcal{Y}_L|-d_L-1)}$ decay exponentially fast to zero as the cluster size $|\mathcal{Y}_L|$ grows, therefore negligible. Accordingly, when the subspaces are well separated from each other and the data size is large enough, GOMP can notch up higher true neighbor recovery rate as compared to OMP.

The above two theorems specify the true neighbor recovery rate up to the $\lceil d_L/p\rceil$th iteration, whence the GOMP is about to leave off as argued in Section II-B. The next theorem then confirms such a stopping condition is valid with a high probability.

***Theorem 3.3:*** Under the same setup as in Theorem 3.1, GOMP in conjunction with the proposed stopping rule halts till the $\lceil d_L/p\rceil$th iteration with a probability exceeding

$$1-Ne^{-n/8}-v(u)\left(\frac{\sigma}{\sqrt{\pi}}\right)^{u}-2pe^{-\sqrt{n/p}}-\left\lceil\frac{d_L}{p}\right\rceil\left[\frac{2e(N-|\mathcal{Y}_L|)}{N^{8\log N/d_L}}+\left(\sqrt{\frac{2}{\pi}}\tau\right)^{(|\mathcal{Y}_L|-d_L-p)}\left(\frac{e(|\mathcal{Y}_L|-1)}{p-1}\right)^{p-1}+\frac{4+2c}{N^2}\right]. \tag{3.9}$$

where $u\triangleq\min_{r\in\mathbb{Z},\ pr<d_L}\left(d_L-pr\right)$ and $v(u)$ is the volume of unit-ball in $\mathbb{R}^u$.

*[Proof]:* See Section V-C. □

Under Assumptions 1 to 3, Theorem 3.3 provides an analytic probability lower bound for the event that GOMP in the high-SNR regime successfully recovers about as many neighbors as the subspace dimension $d_L$. We remark that, when the noise power increases further, the dimension of the span of data points in $\mathcal{Y}_L$ would exceed $d_L$; accordingly, the algorithm would stop till more than $\lceil d_L / p \rceil$ iterations in order to recover more than $d_L$ neighbors. As the noise power grows higher, the residual is soon dominated by noise to in turn trigger the proposed stopping rule (2.3), rendering the algorithm terminated in fewer than $\lceil d_L / p \rceil$ iterations with scant neighbors. This will be seen in the simulation section.

# IV. EXPERIMENTAL RESULTS

In this section, numerical simulations based on both synthetic data and real human face data are used to corroborate our theoretical study and illustrate the performance of the proposed GOMP method. Synthetic data are generated similar to [7, 14]. The ground truth is a union of three subspaces $\mathcal{S}_1$, $\mathcal{S}_2$ and $\mathcal{S}_3$ of $\mathbb{R}^{350}$, with an equal subspace dimension $d_1 = d_2 = d_3 = 6$; separation between two distinct subspaces is gauged by the subspace affinity defined in (3.2). Noiseless signal points are randomly and uniformly drawn from each subspace, and are corrupted by Gaussian noise with zero mean and standard deviation $\sigma$. The sampling density is defined to be $\phi_k \triangleq |\mathcal{Y}_k| / d_k$, $1 \le k \le 3$. Regarding the neighbor identification performance metrics, we consider the *true neighbor rate* (TNR) [14], that is,

$$TNR \triangleq \frac{\sum_{(i,j)\in\mathcal{T}} \mathbf{1}(c_{i,j}^* \neq 0)}{\sum_{1\le i,j\le N} \mathbf{1}(c_{i,j}^* \neq 0)}, \quad \mathcal{T} \triangleq \{(i,j) |\ \mathbf{y}_i, \mathbf{y}_j \in \mathcal{Y}_l, \text{ for some } 1 \le l \le L\} \tag{4.1}$$

where $c_{i,j}^*$ is the $j$th entry of the computed sparse representation vector $\mathbf{c}_i^*$ using GOMP/OMP (see step 4 of the algorithm), and the *average number of recovered neighbors* (ANRN)

$$ANRN \triangleq \frac{1}{N}\sum_{i=1}^{N} \|\mathbf{c}_i^*\|_0, \tag{4.2}$$

which assesses the ability to boost neighbor acquisition. As in [12-14], the global clustering performance is evaluated on the basis of the *correct clustering rate* (CCR), defined as

$$CCR \triangleq (\#\ \text{of correctly clustered data points}) / N. \tag{4.3}$$

## *A. Synthetic Data*

We first use synthetic data to test the performance of the proposed method. To ease illustration, we consider the case that $aff(\mathcal{S}_k, \mathcal{S}_l) = \rho$ for all $1 \le k \neq l \le 3$, i.e., the three subspaces are equally separated from each other, and that $\phi_1 = \phi_2 = \phi_3 = \phi$, so with equal sample density. Associated with $9$ different pairs $(\rho, \phi)$ of subspace affinity and sample density, Fig. 2 and 3 plot the simulated TNR and CCR, respectively, versus noise standard deviation $\sigma$; sub-figures on the same row (column, respectively) correspond to an identical $\phi$ ($\rho$, respectively), while $\rho$ ($\phi$, respectively) is increased when going from left to right (top to bottom, respectively). In generating Fig. 2 and 3, the total number of recovered neighbors is fixed to be 6 (equal to the subspace dimension), and the total number of iterations is pre-set to be $M = 6$ for OMP and $M = \lceil 6/p \rceil$ for GOMP, respectively; this represents the ideal situation that subspace dimension is perfectly known. The figures show that, as compared to the conventional OMP ($p = 1$), the proposed GOMP with multiple neighbor selection in all cases achieves higher TNR, and in turn brings about higher CCR. The above experiment is carried out again by instead using the proposed stopping rule (2.3) to halt the algorithm, and the results are shown in Fig. 4 and Fig. 5.

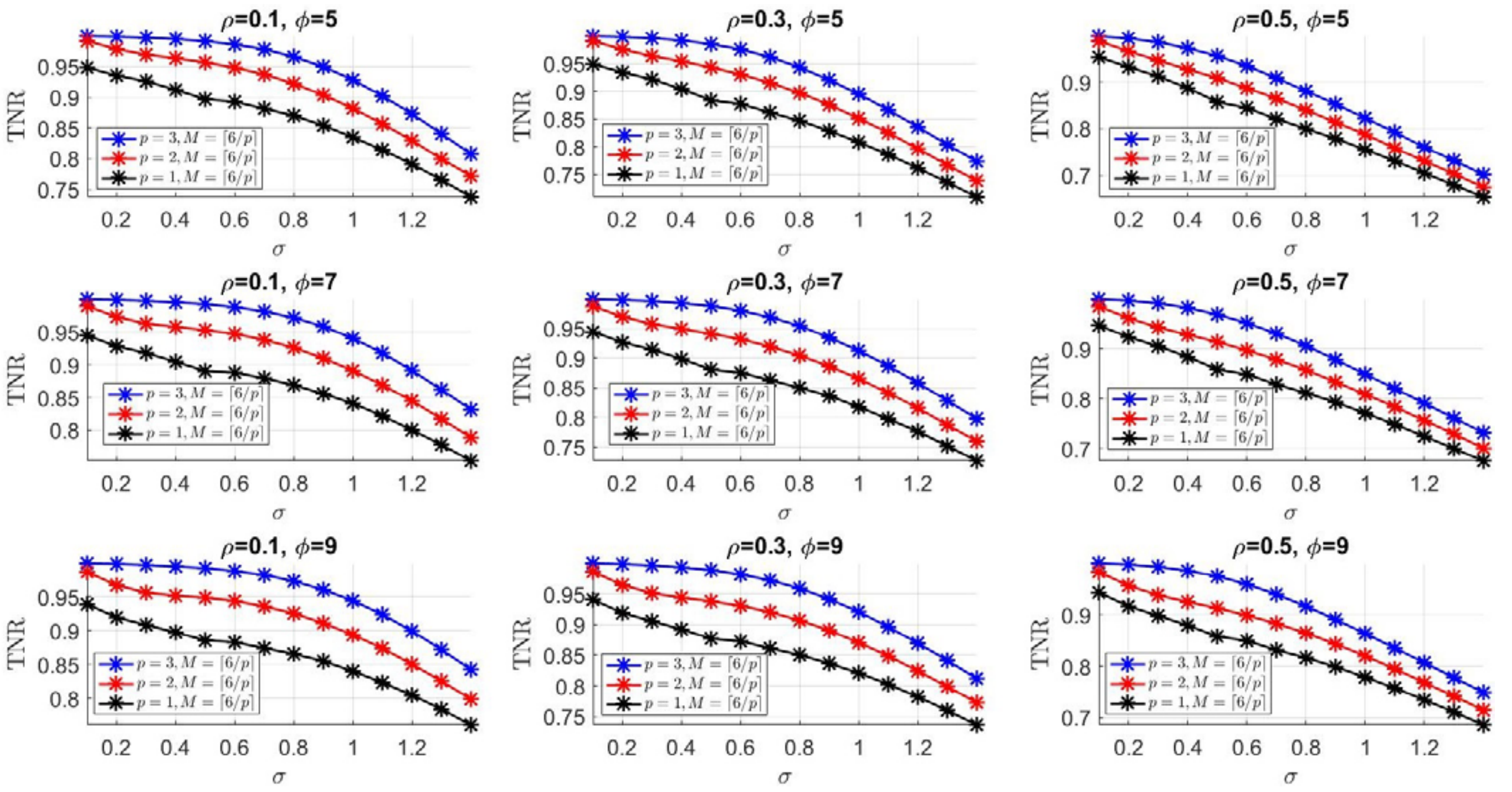

Fig. 2. TNR versus noise standard deviation $\sigma$ for nine different pairs of subspace affinity $\rho$ and sample density $\phi$; assume the subspace dimension is known beforehand and the number of iterations is preset to be $M = \lceil 6/p \rceil$ ( $p = 1,2,3$ ).

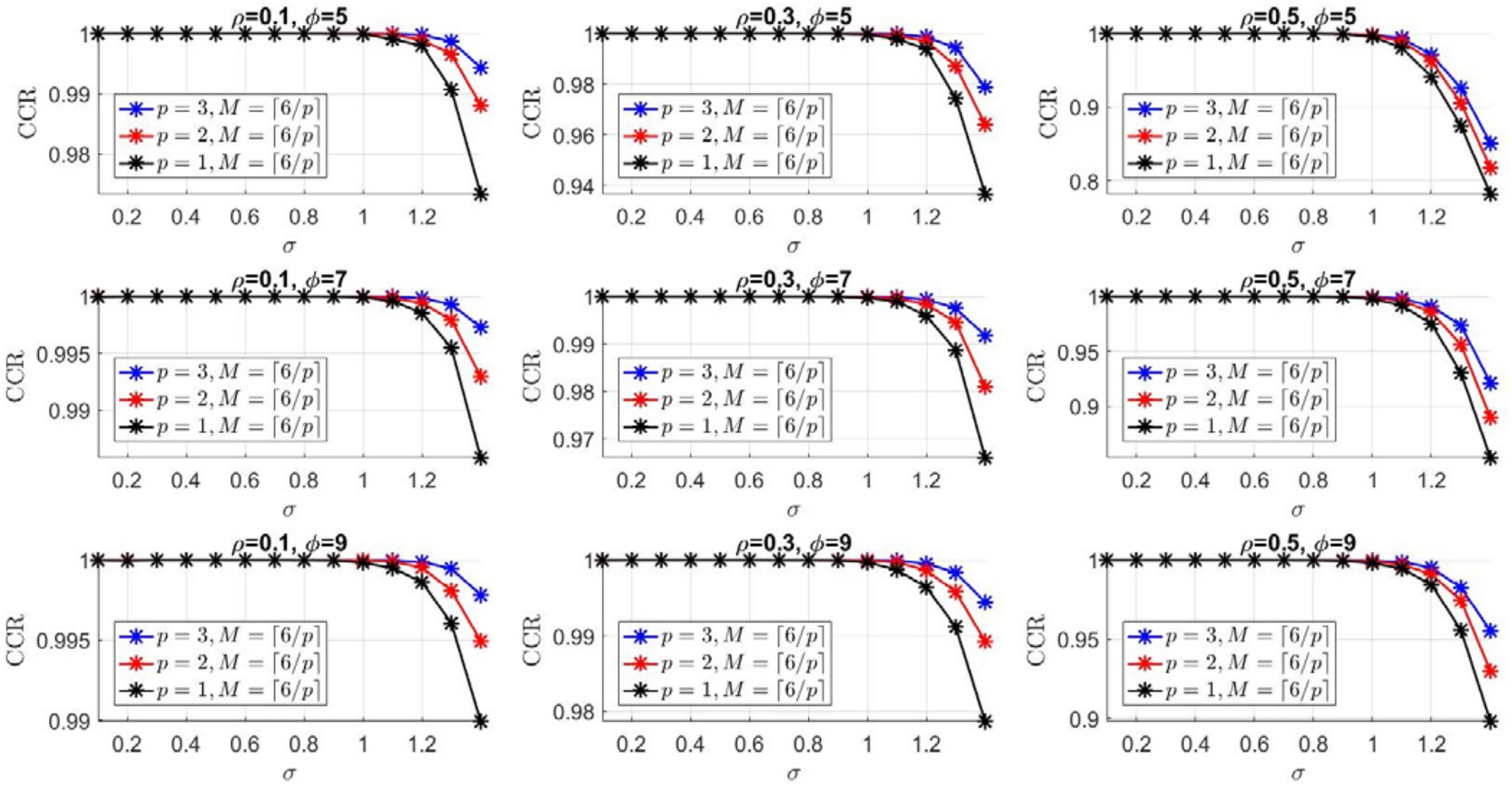

Fig. 3. CCR versus noise standard deviation $\sigma$ for nine different pairs of subspace affinity $\rho$ and sample density $\phi$; assume the subspace dimension is known beforehand and the number of iterations is preset to be $M = \lceil 6/p \rceil$ ( $p = 1,2,3$ ).

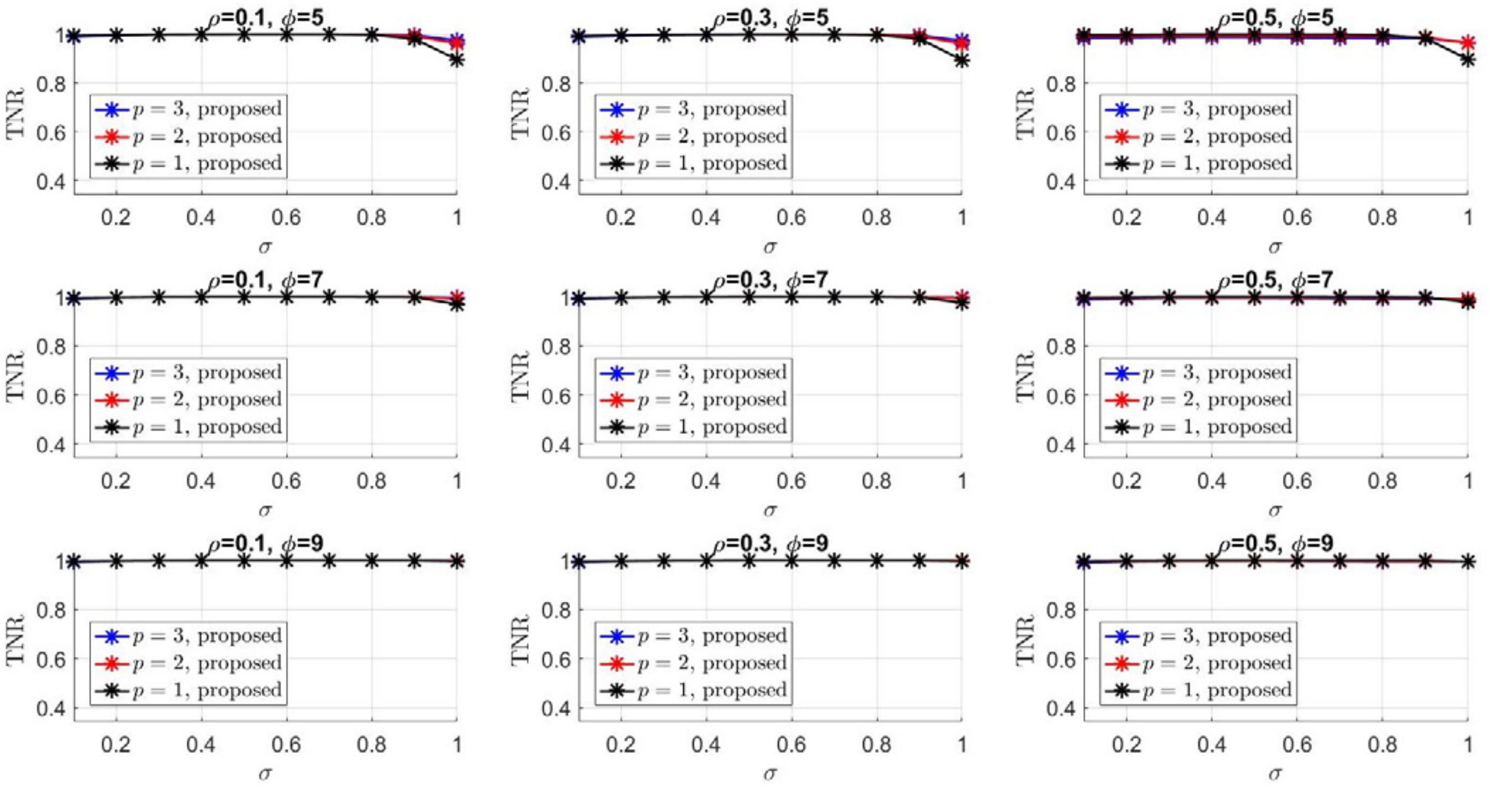

Fig. 4. TNR versus noise standard deviation $\sigma$ for nine different pairs of subspace affinity $\rho$ and sample density $\phi$; the proposed stopping rule (2.3) is used.

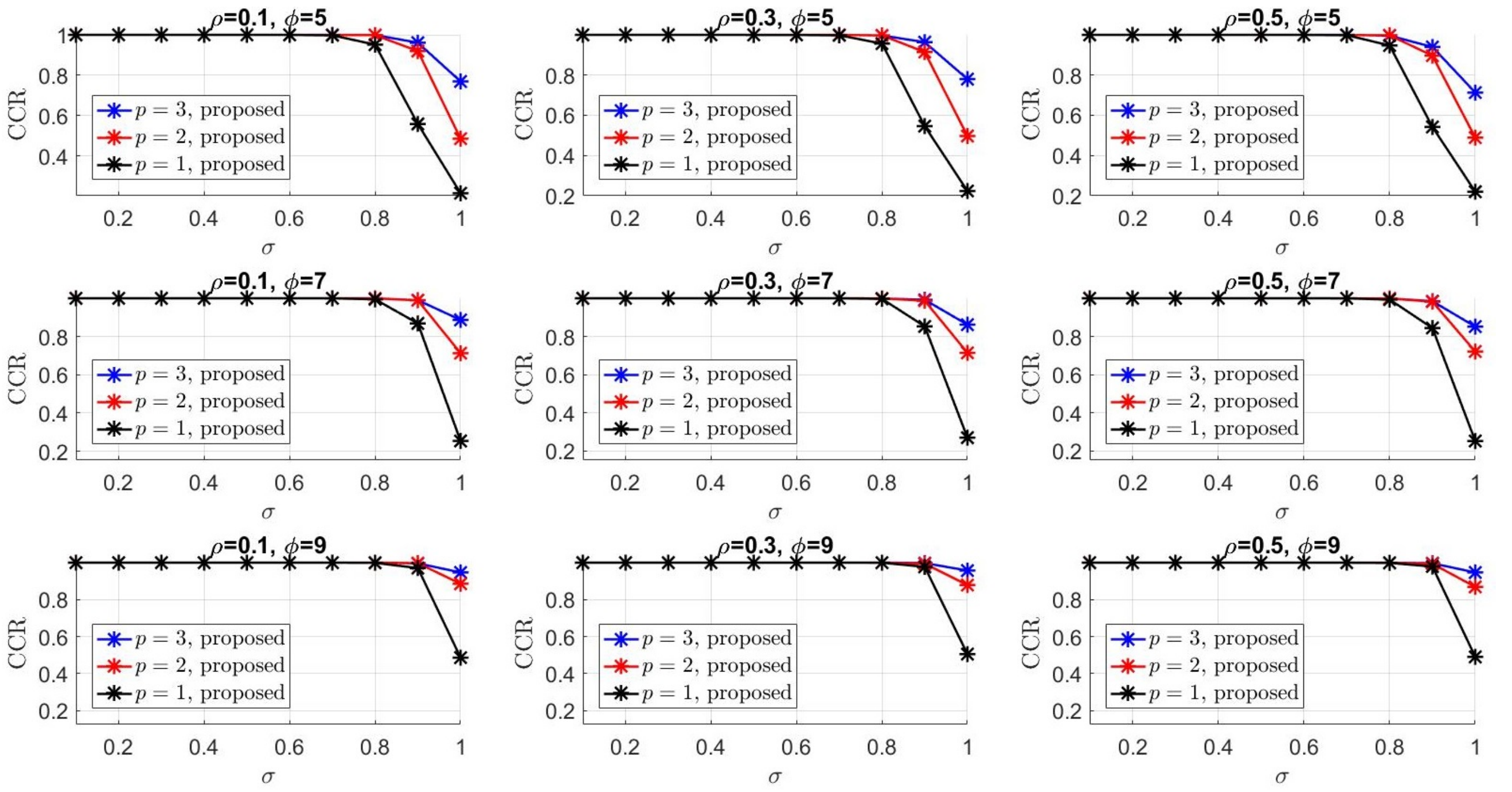

Fig. 5. CCR versus noise standard deviation $\sigma$ for nine different pairs of subspace affinity $\rho$ and sample density $\phi$; the proposed stopping rule (2.3) is used.

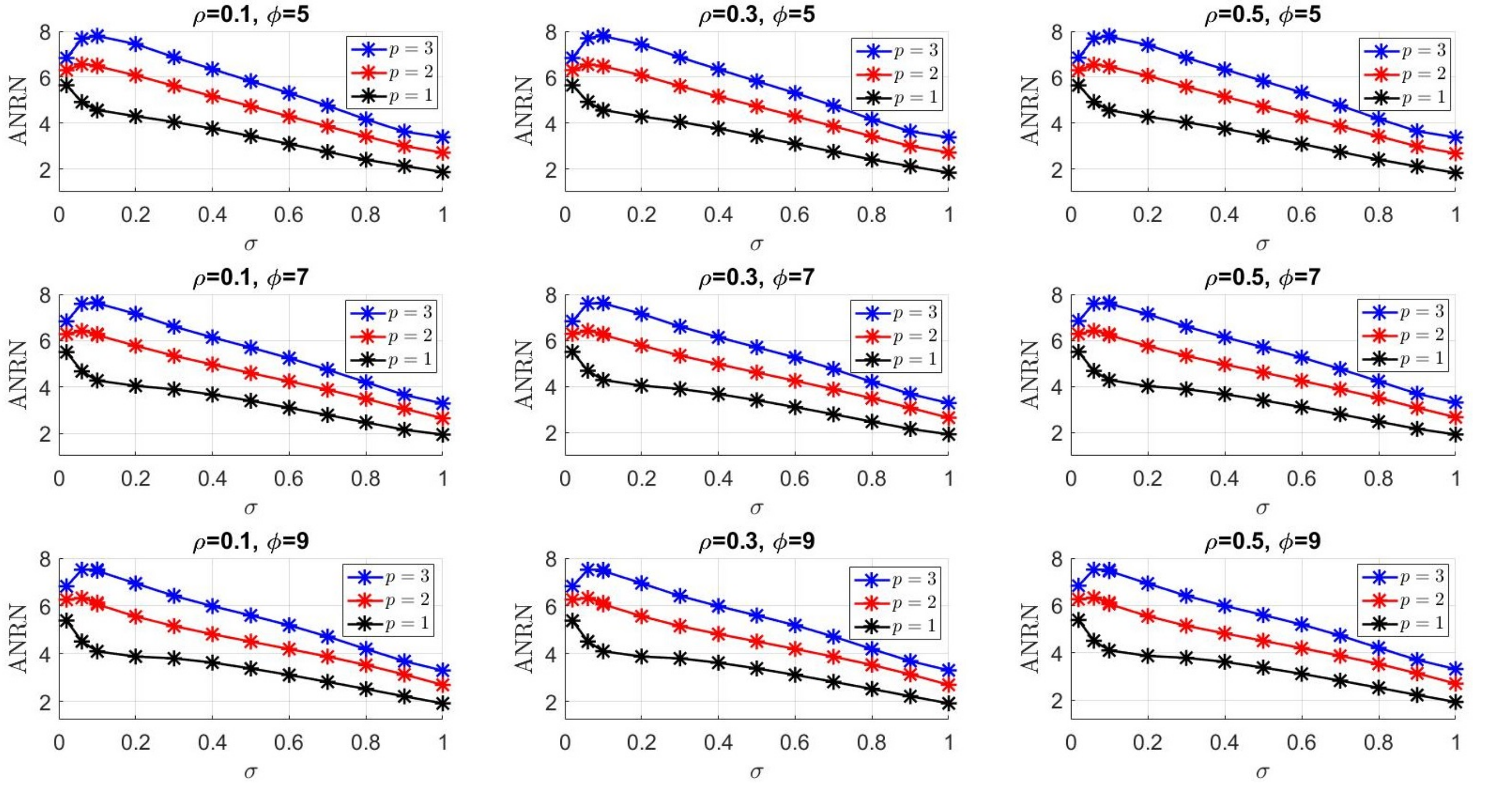

Fig. 6. ANRN versus noise standard deviation $\sigma$ for nine different pairs of subspace affinity $\rho$ and sample density $\phi$; the proposed stopping rule (2.3) is used.

It can be seen that the proposed GOMP still outperforms OMP. Fig. 6 further plots the ANRN of the two methods, both employing the proposed stopping rule (2.3), for different noise standard deviation $\sigma$. We first observe from the figure that, when noise is so small that $\sigma < 0.1$, ANRN of all cases is about six, in support of the assertion in Theorem 3.3 that as many neighbors as the subspace dimension suffice to well explain the data point under consideration. As $\sigma$ grows up to $0.1$, the proposed GOMP (with $p = 2, 3$) tends to retrieve more neighbors. As we have mentioned in the discussions at the end of Section III, further noise corruption would enlarge the dimension of the span of the data cluster; as a result, the algorithm is terminated after more than $\lceil d_L / p \rceil$ iterations so as to recover more neighbors (than subspace dimension) to explain the data point. When $\sigma$ gets even larger, the residual is then severely dominated by noise, only to halt neighbor search earlier, say, in less than $\lceil d_L / p \rceil$ iterations, ending up with fewer recovered neighbors and reduced ANRN.

## *B. Real Human Face Data*

We proceed to test the performance of the proposed method using the Extended Yale B human face dataset [7], comprised of photos of $38$ people each with $65$ images of $192\text{x}168$ pixels. To reduce computations, the same dimensionality reduction technique as in [7] is employed for reducing the image size to $48\text{x}42$. To determine the subspace dimensions, associated with each individual (cluster) we vectorize all $65$ dimensionality-reduced images, stack them into a matrix, and calculate the normalized singular values (with respective to the maximal one). The results are plotted in Fig. 7 for 7 randomly chosen individuals. It is seen that all curves have a knee around $7 \sim 9$. Our further simulation study based on many different sets of individuals also demonstrates similar phenomenon, indicating that data vectors from each cluster lie in a low-dimensional subspace with dimension nearly $7 \sim 9$, in our simulations below simply set to be $8$, the average.

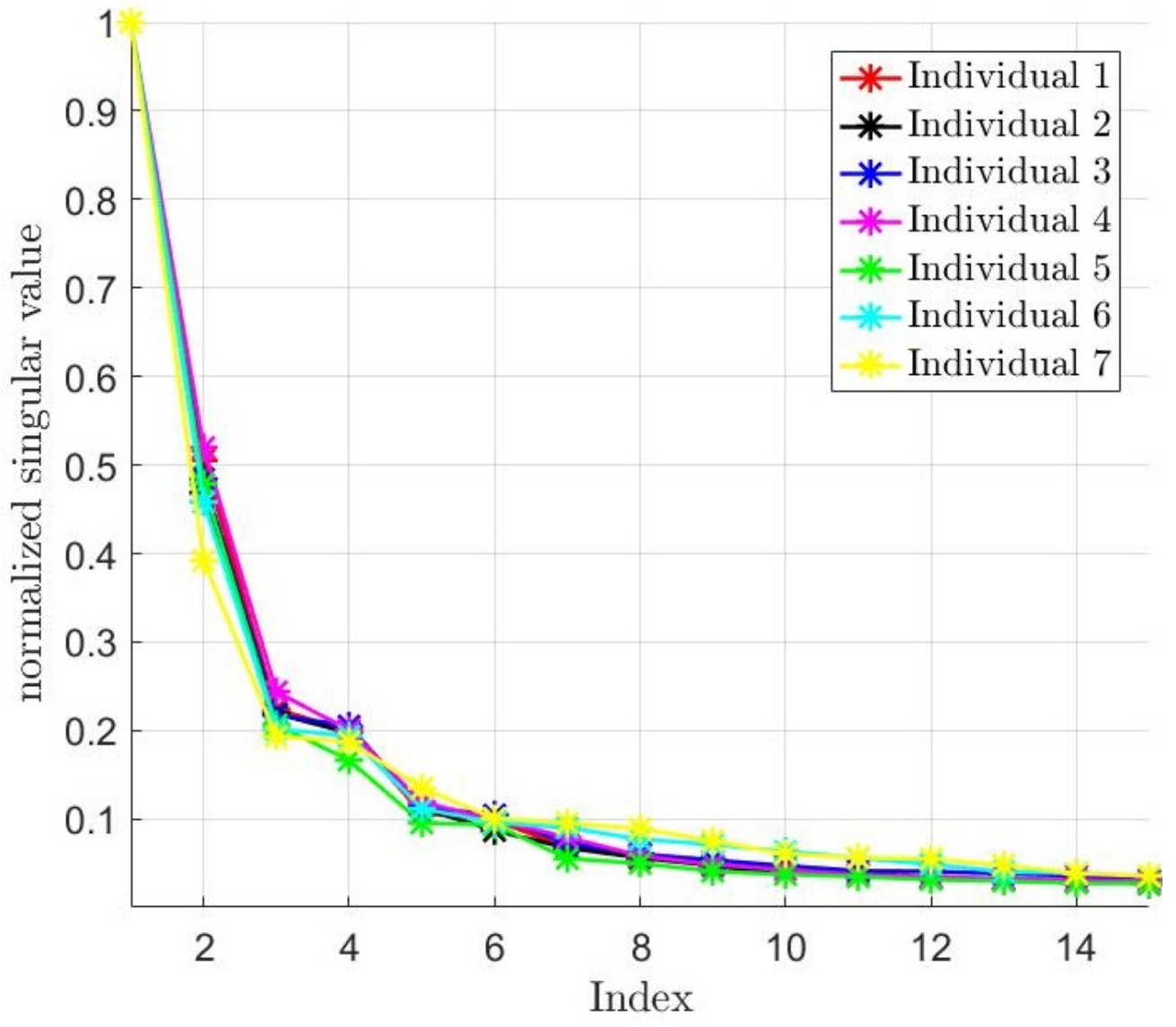

Fig. 7. Normalized singular values of 7 randomly chosen individual data matrices of Extended Yale B human face dataset.

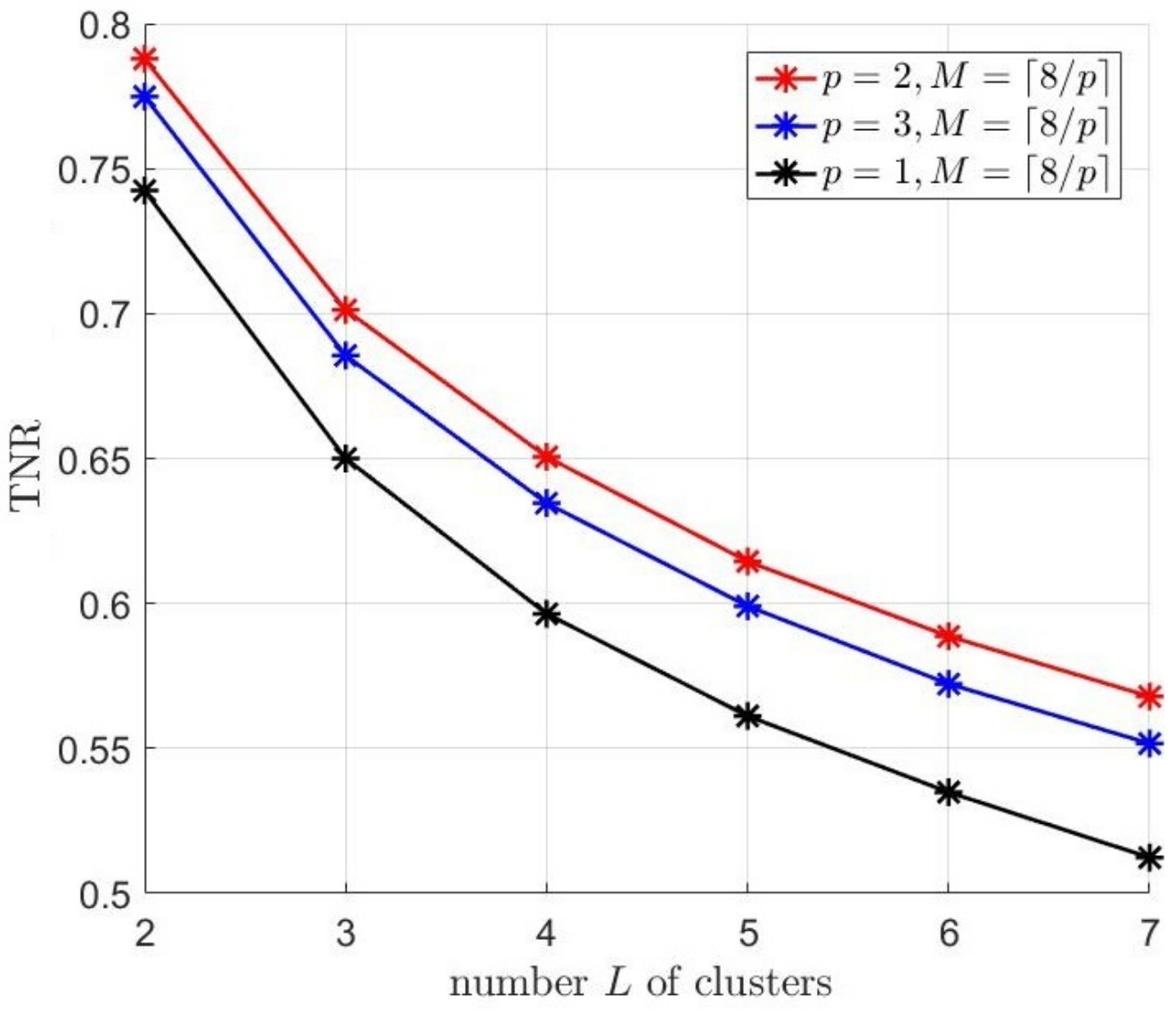

Fig. 8. TNR versus number $L$ of clusters for Extended Yale B human face dataset; number of iterations fixed to $M = \lceil 8/p \rceil$.

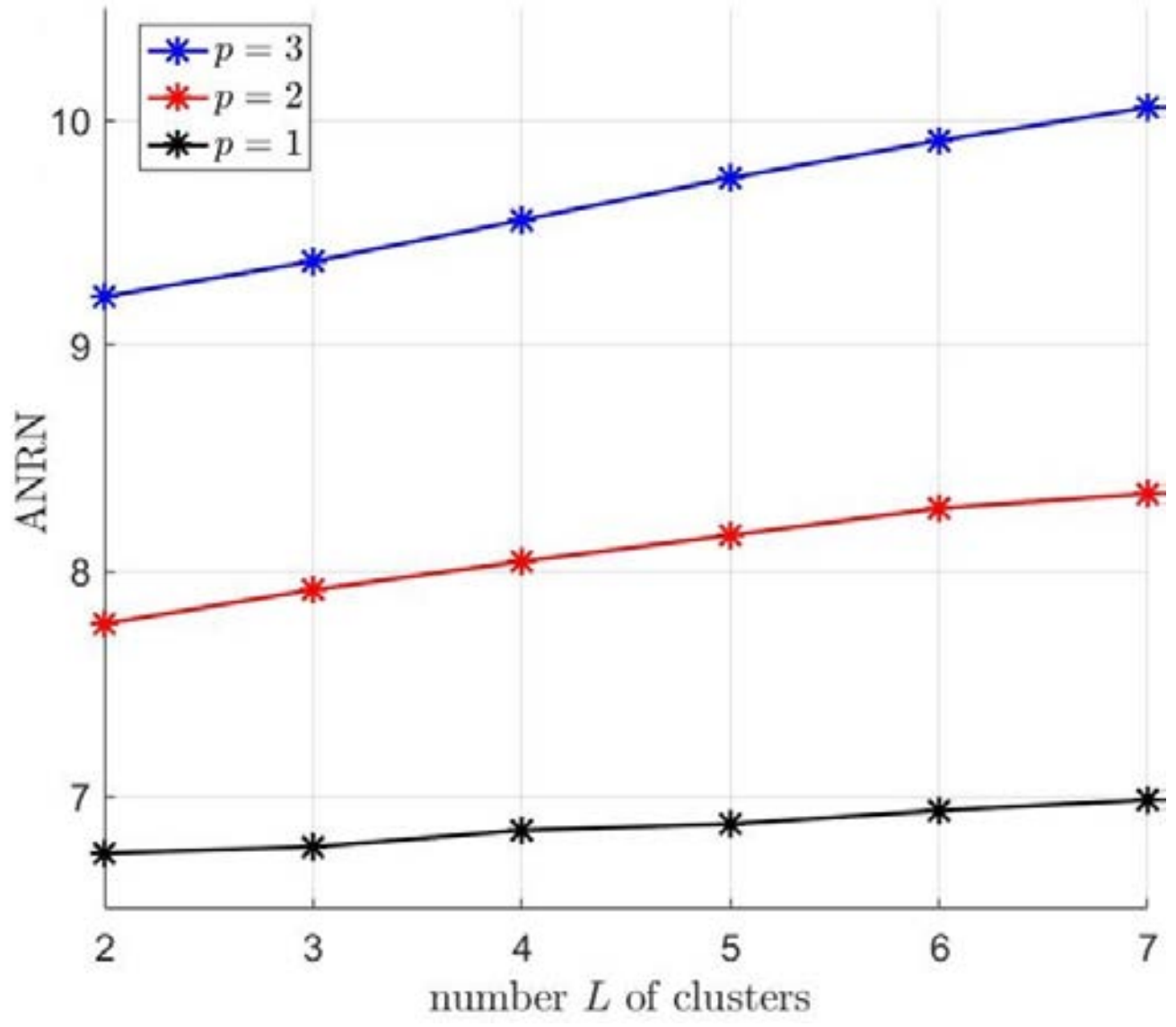

Fig. 9. ANRN versus number $L$ of clusters for Extended Yale B human face dataset; the proposed stopping rule (2.3) is used.

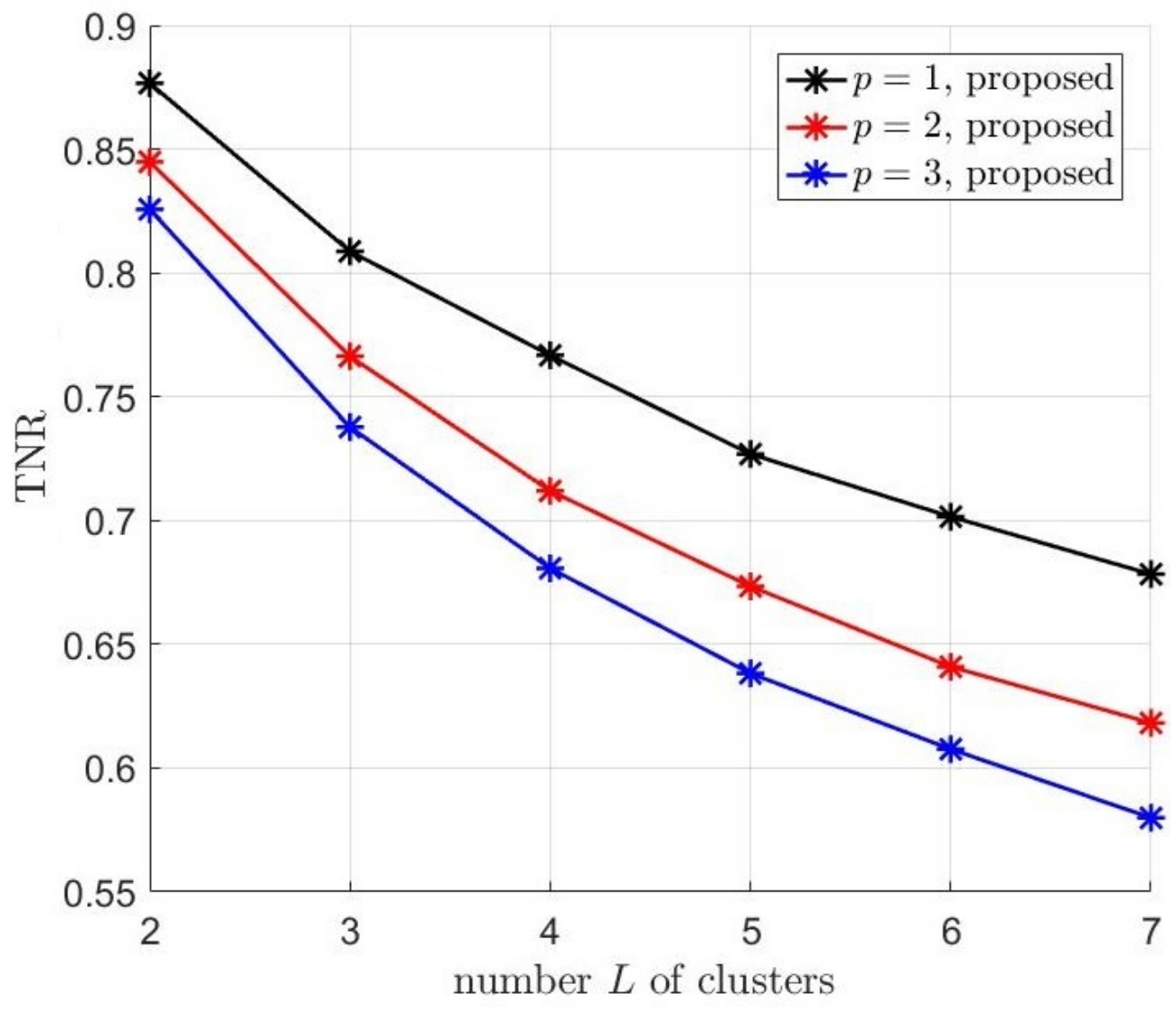

Fig. 10. TNR versus number $L$ of clusters for Extended Yale B human face dataset; the proposed stopping rule (2.3) is used.

Assuming perfect knowledge of subspace dimension, Fig. 8 compares TNR for different number $L$ of clusters; both GOMP and OMP are halted once the number of iterations reaches $M = \lceil 8/p \rceil$ (for $p = 1,2,3$, the numbers of recovered neighbors are $8$, $8$, and $9$, respectively). The figure shows that, as expected, the proposed GOMP outperforms OMP. Fig. 9 compares ANRN when the proposed stopping rule (2.3) is employed. We can first see from the figure that ANRN assumes values around $7 \sim 9$ (subspace dimensions), in agreement with the assertion in Theorem 3.3. Then, similar to the synthetic data case (in Fig. 6), OMP is seen to yield lowest ANRN, mainly because its residual is soon dominated by noise to trigger the stopping condition (2.3), as we have explained in the previous subsection. Lastly, ANRN is observed to increase with $L$; the reason behind is that, when more clusters are present (large $L$), the data set gets more crowded, the algorithm then apt to mis-identify some neighbors from other clusters, ending up with more total recovered neighbors to ensure enough true. Fig. 10 goes on to compare the corresponding TNR, showing that OMP yields highest TNR, due to its low ANRN. Finally, Fig. 11 plots the resultant CCR; the genie-added case with perfect knowledge of subspace dimension (i.e., pre-set $M = \lceil 8/p \rceil$ iterations) is also included as the benchmark. The figure shows that the proposed GOMP outperforms conventional OMP, thanks to higher ANRN and therefore better graph connectivity. In addition, GOMP combined with the proposed stopping rule (2.3), while dispensing with subspace dimension estimation, compares favorably with the genie-aided benchmark, once again confirming the merits of the proposed scheme.

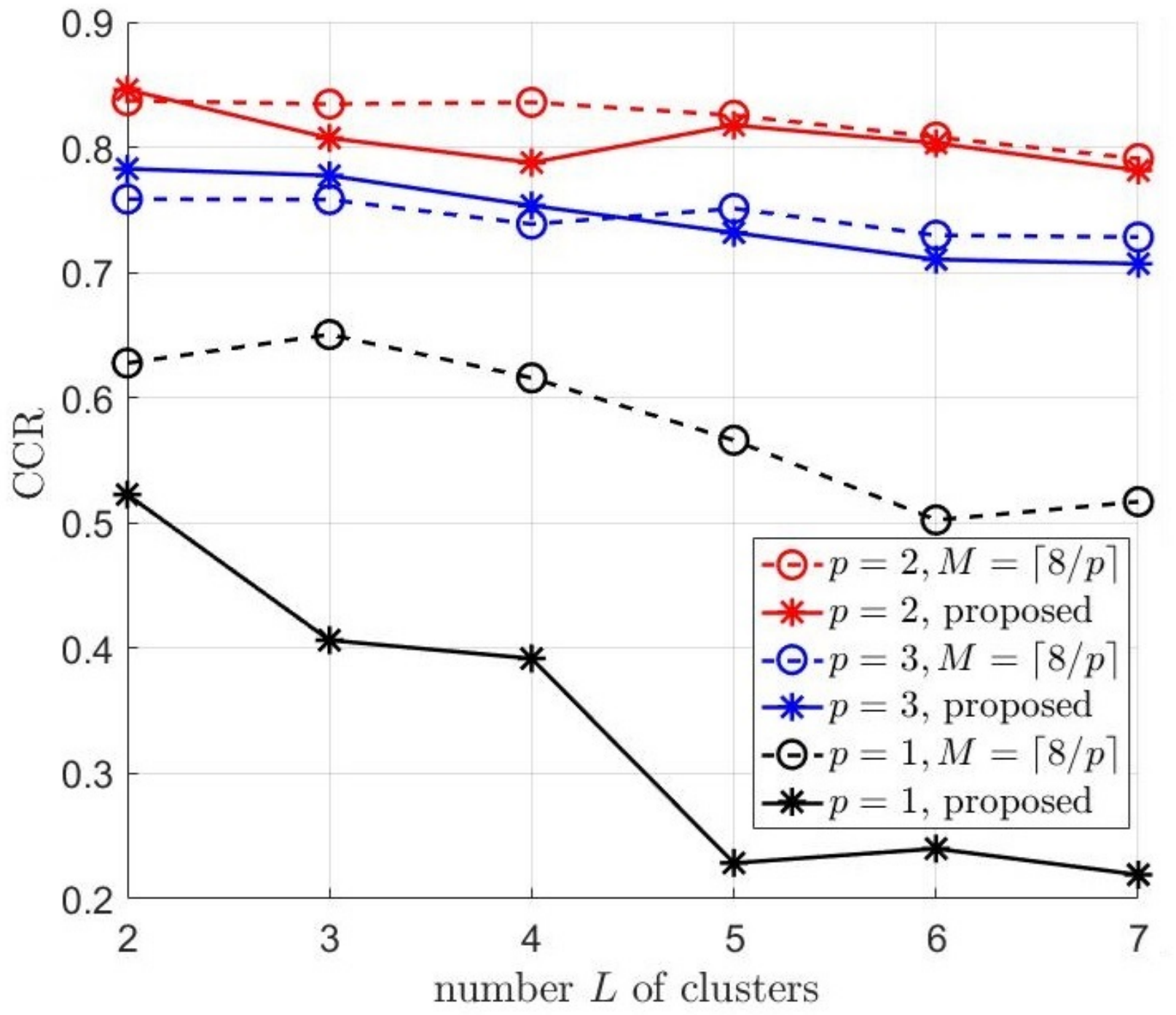

Fig. 11. CCR versus number $L$ of clusters for Extended Yale B human face dataset; the proposed stopping rule (2.3) is used.

# V. PROOFS

## *A. Proof of Theorem 3.1*

We set about the proof by defining the following per-iteration neighbor recovery event:

$$E_m \triangleq \{\text{at least } k_m \text{ true neighbors are obtained in the } m\text{th iteration}\}, \quad 1 \le m \le M. \tag{5.1}$$

It suffices to show the event $\bigcap_{m=1}^{M} E_m$ holds with a probability as high as claimed in (3.4). The following lemmas are needed for deriving Theorem 3.1.

***Lemma 5.1:*** Let $\mathbf{a} \in \mathbb{R}^m$ be uniformly distributed over the unit-sphere of $\mathbb{R}^m$, and $\mathbf{b} \in \mathbb{R}^m$ be a random vector independent of $\mathbf{a}$. Then, for $\varepsilon \geq 0$ we have

(a) $\Pr\left\{\left|\mathbf{a}^T\mathbf{b}\right| > \varepsilon\left\|\mathbf{b}\right\|_2\right\} \leq 2e^{-m\varepsilon^2/2}$, (5.2)

(b) $\Pr\left\{\left|\mathbf{a}^T\mathbf{b}\right| < \varepsilon\left\|\mathbf{b}\right\|_2 / \sqrt{m}\right\} \leq \sqrt{2/\pi}\varepsilon$. (5.3)

*[Proof]* : See Appendix A. □

***Lemma 5.2:*** Let $\mathbf{U}_l \in \mathbb{R}^{n \times d_l}$ be a matrix whose columns form an orthonormal basis for subspace $\mathcal{S}_l$, $1 \leq l \leq L$. For $\mathbf{x}_i \notin \mathcal{S}_L$, we have

$$\Pr\left\{\left|\left\langle \mathbf{x}_i, \frac{\mathbf{r}_{m,\|}^{(N)}}{\left\|\mathbf{r}_{m,\|}^{(N)}\right\|_2}\right\rangle\right| > \max_{l \neq L} 4\log(N)\frac{\left\|\mathbf{U}_l^T\mathbf{U}_L\right\|_F}{\sqrt{d_l d_L}}\right\} < \frac{2}{N^{(8\log N)/d_L}}. \quad (5.4)$$

*[Proof]* : See Appendix B. □

***Lemma 5.3:*** For $\mathbf{x}_i \notin \mathcal{S}_L$, we have

$$\Pr\left\{\left|\left\langle \mathbf{x}_i, \frac{\mathbf{r}_{m,\perp}^{(N)}}{\left\|\mathbf{r}_{m,\perp}^{(N)}\right\|_2}\right\rangle\right| \leq \sqrt{\frac{6\log N}{n-d_L}}\right\} \geq 1 - \frac{2c}{N^3} \text{ for some constant } c > 0. \quad (5.5)$$

*[Proof]* : See Appendix C. □

If $k_m = 0$, it follows $\Pr\{E_m\} = 1$ because the number of true neighbors is never negative. We then consider the case that $k_m > 0$. Since the already chosen data vectors in all the previous $m-1$ iterations are orthogonal to the residual $\mathbf{r}_{m-1}^{(N)}$, i.e., $\left|\left\langle \mathbf{y}_j, \mathbf{r}_{m-1}^{(N)}\right\rangle\right| = 0$ for all $j \in \Lambda_{m-1}$, the "yet-to-be-selected" candidate neighbors are $\mathbf{y}_j$ 's for $j \in \{1,2,\cdots,N-1\} \setminus \Lambda_{m-1}$. Given that $\Lambda_{m-1} = \Lambda_{m-1}^t \cup \Lambda_{m-1}^f$, where the disjoint subsets $\Lambda_{m-1}^t$ and $\Lambda_{m-1}^f$ consist of, respectively, the indexes of true and false neighbors already selected up to the first $m-1$ iterations, $|\mathcal{Y}_L| - 1 - \left|\Lambda_{m-1}^t\right|$ and $N - |\mathcal{Y}_L| - \left|\Lambda_{m-1}^f\right|$ data vectors remain in the same and different clusters as $\mathbf{y}_N$, respectively. Let the true neighbor candidates $\mathbf{y}_{t_j}$, where $\mathbf{x}_{t_j} \in \mathcal{S}_L$ and $t_j \notin \Lambda_{m-1}^t$, be sorted according to $\left|\left\langle \mathbf{y}_{t_1}, \mathbf{r}_{m-1}^{(N)}\right\rangle\right| \geq \left|\left\langle \mathbf{y}_{t_2}, \mathbf{r}_{m-1}^{(N)}\right\rangle\right| \geq \cdots \geq \left|\left\langle \mathbf{y}_{t_{|\mathcal{Y}_L|-1-\left|\Lambda_{m-1}^t\right|}}, \mathbf{r}_{m-1}^{(N)}\right\rangle\right|$, and likewise the false neighbor candidates $\mathbf{y}_{f_k}$, where $\mathbf{x}_{f_k} \notin \mathcal{S}_L$ and $f_k \notin \Lambda_{m-1}^f$, in the way that $\left|\left\langle \mathbf{y}_{f_1}, \mathbf{r}_{m-1}^{(N)}\right\rangle\right| \geq \left|\left\langle \mathbf{y}_{f_2}, \mathbf{r}_{m-1}^{(N)}\right\rangle\right| \geq \cdots \geq \left|\left\langle \mathbf{y}_{f_{N-|\mathcal{Y}_L|-\left|\Lambda_{m-1}^f\right|}}, \mathbf{r}_{m-1}^{(N)}\right\rangle\right|$. Also, let the noiseless signal vectors $\mathbf{x}_{\bar{t}_g} \in \mathcal{S}_L$, $\bar{t}_g \notin \Lambda_{m-1}^t$, be ordered so that $\left|\left\langle \mathbf{x}_{\bar{t}_1}, \mathbf{r}_{m-1,\|}^{(N)}\right\rangle\right| \geq \left|\left\langle \mathbf{x}_{\bar{t}_2}, \mathbf{r}_{m-1,\|}^{(N)}\right\rangle\right| \geq \cdots \geq \left|\left\langle \mathbf{x}_{\bar{t}_{|\mathcal{Y}_L|-1-\left|\Lambda_{m-1}^t\right|}}, \mathbf{r}_{m-1,\|}^{(N)}\right\rangle\right|$, while those $\mathbf{x}_{\bar{f}_h} \notin \mathcal{S}_L$, $\bar{f}_h \notin \Lambda_{m-1}^f$, in the way $\left|\left\langle \mathbf{x}_{\bar{f}_1}, \mathbf{r}_{m-1,\|}^{(N)}\right\rangle\right| \geq \left|\left\langle \mathbf{x}_{\bar{f}_2}, \mathbf{r}_{m-1,\|}^{(N)}\right\rangle\right| \geq \cdots \geq \left|\left\langle \mathbf{x}_{\bar{f}_{N-|\mathcal{Y}_L|-\left|\Lambda_{m-1}^f\right|}}, \mathbf{r}_{m-1,\|}^{(N)}\right\rangle\right|$.

The event $E_m$ defined in (5.1) can be expressed in terms of the above ordered statistics. By definition,

$E_m$ occurs when at least $k_m$ true neighbors, and so at most $p-k_m$ false neighbors, are recovered in the $m$th iteration. That is to say, $\mathbf{y}_{t_{k_m}}$ is chosen as a neighbor but $\mathbf{y}_{f_{p-k_m+1}}$ is not, justifying the inequality $\left|\left\langle \mathbf{y}_{f_{p-k_m+1}}, \mathbf{r}_{m-1}^{(N)}\right\rangle\right| \le \left|\left\langle \mathbf{y}_{t_{k_m}}, \mathbf{r}_{m-1}^{(N)}\right\rangle\right|$. The converse is obviously true. Hence we can rewrite $E_m$ as

$$E_m = \left\{\left|\left\langle \mathbf{y}_{f_{p-k_m+1}}, \mathbf{r}_{m-1}^{(N)}\right\rangle\right| \le \left|\left\langle \mathbf{y}_{t_{k_m}}, \mathbf{r}_{m-1}^{(N)}\right\rangle\right|\right\}. \tag{5.6}$$

Now the expression (5.6) involves ordered statistics of the absolute inner products between the residual and noisy data points $\mathbf{y}_j$'s, whose distribution is however quite complicated. To ease our lower bound derivation, we shall instead seek sufficient conditions, specified by absolute inner products between the residual and *noise-free* signal points $\mathbf{x}_j$'s, for $\left|\left\langle \mathbf{y}_{f_{p-k_m+1}}, \mathbf{r}_{m-1}^{(N)}\right\rangle\right| \le \left|\left\langle \mathbf{y}_{t_{k_m}}, \mathbf{r}_{m-1}^{(N)}\right\rangle\right|$; this allows us to employ the assumed uniform distribution of $\mathbf{x}_j$'s (Assumption 1) as well as the ground-truth subspace orientation (Assumption 3) to facilitate analysis. For this we first note from (3.1) that, for all $1 \le j \le N-1$,

$$\begin{aligned}\left\langle \mathbf{y}_j, \mathbf{r}_{m-1}^{(N)}\right\rangle &= \left\langle \mathbf{x}_j + \mathbf{e}_j, \mathbf{r}_{m-1}^{(N)}\right\rangle \\ &= \left\langle \mathbf{x}_j, \mathbf{r}_{m-1}^{(N)}\right\rangle + \left\langle \mathbf{e}_j, \mathbf{r}_{m-1}^{(N)}\right\rangle \\ &= \left\langle \mathbf{x}_j, \mathbf{r}_{m-1,\|}^{(N)}\right\rangle + \left\langle \mathbf{x}_j, \mathbf{r}_{m-1,\perp}^{(N)}\right\rangle + \left\langle \mathbf{e}_j, \mathbf{r}_{m-1}^{(N)}\right\rangle.\end{aligned} \tag{5.7}$$

Then, for all $u \le k_m$, it follows

$$\begin{aligned}\left|\left\langle \mathbf{y}_{\bar{t}_u}, \mathbf{r}_{m-1}^{(N)}\right\rangle\right| &= \left|\left\langle \mathbf{x}_{\bar{t}_u}, \mathbf{r}_{m-1,\|}^{(N)}\right\rangle + \left\langle \mathbf{x}_{\bar{t}_u}, \mathbf{r}_{m-1,\perp}^{(N)}\right\rangle + \left\langle \mathbf{e}_{\bar{t}_u}, \mathbf{r}_{m-1}^{(N)}\right\rangle\right| \\ &\ge \left|\left\langle \mathbf{x}_{\bar{t}_u}, \mathbf{r}_{m-1,\|}^{(N)}\right\rangle\right| - \max_{1\le j\le|\mathcal{Y}_L|-1-\left|\Lambda_{m-1}^t\right|}\left|\left\langle \mathbf{x}_{t_j}, \mathbf{r}_{m-1,\perp}^{(N)}\right\rangle\right| - \max_{1\le j\le|\mathcal{Y}_L|-1-\left|\Lambda_{m-1}^t\right|}\left|\left\langle \mathbf{e}_{t_j}, \mathbf{r}_{m-1}^{(N)}\right\rangle\right| \\ &\overset{(a)}{\ge} \left|\left\langle \mathbf{x}_{\bar{t}_u}, \mathbf{r}_{m-1,\|}^{(N)}\right\rangle\right| - \max_{1\le j\le|\mathcal{Y}_L|-1-\left|\Lambda_{m-1}^t\right|}\left|\left\langle \mathbf{e}_{t_j}, \mathbf{r}_{m-1}^{(N)}\right\rangle\right| \\ &\overset{(b)}{\ge} \left|\left\langle \mathbf{x}_{\bar{t}_{k_m}}, \mathbf{r}_{m-1,\|}^{(N)}\right\rangle\right| - \max_{1\le j\le|\mathcal{Y}_L|-1-\left|\Lambda_{m-1}^t\right|}\left|\left\langle \mathbf{e}_{t_j}, \mathbf{r}_{m-1}^{(N)}\right\rangle\right|,\end{aligned} \tag{5.8}$$

where (a) holds since $\left\langle \mathbf{x}_{t_j}, \mathbf{r}_{m-1,\perp}^{(N)}\right\rangle = 0$ for $\mathbf{x}_{t_j} \in \mathcal{S}_L$, and (b) follows from the ordering of $\left|\left\langle \mathbf{x}_{\bar{t}_g}, \mathbf{r}_{m-1,\|}^{(N)}\right\rangle\right|$'s. Hence, at least $k_m$ absolute inner products $\left|\left\langle \mathbf{y}_{\bar{t}_u}, \mathbf{r}_{m-1}^{(N)}\right\rangle\right|$ are no smaller than the term on the right-hand-side (RHS) of (5.8); in particular, for $\left|\left\langle \mathbf{y}_{t_{k_m}}, \mathbf{r}_{m-1}^{(N)}\right\rangle\right|$, the $k_m$th largest $\left|\left\langle \mathbf{y}_{t_j}, \mathbf{r}_{m-1}^{(N)}\right\rangle\right|$, we must have

$$\left|\left\langle \mathbf{y}_{t_{k_m}}, \mathbf{r}_{m-1}^{(N)}\right\rangle\right| \ge \left|\left\langle \mathbf{x}_{\bar{t}_{k_m}}, \mathbf{r}_{m-1,\|}^{(N)}\right\rangle\right| - \max_{1\le j\le|\mathcal{Y}_L|-1-\left|\Lambda_{m-1}^t\right|}\left|\left\langle \mathbf{e}_{t_j}, \mathbf{r}_{m-1}^{(N)}\right\rangle\right|. \tag{5.9}$$

Similarly, for all $u \ge p-k_m+1$, it follows *mutatis mutandis* that

$$\begin{aligned}\left|\left\langle \mathbf{y}_{\bar{f}_u}, \mathbf{r}_{m-1}^{(N)}\right\rangle\right| &= \left|\left\langle \mathbf{x}_{\bar{f}_u}, \mathbf{r}_{m-1,\parallel}^{(N)}\right\rangle + \left\langle \mathbf{x}_{\bar{f}_u}, \mathbf{r}_{m-1,\perp}^{(N)}\right\rangle + \left\langle \mathbf{e}_{\bar{f}_u}, \mathbf{r}_{m-1}^{(N)}\right\rangle\right| \\ &\le \left|\left\langle \mathbf{x}_{\bar{f}_u}, \mathbf{r}_{m-1,\parallel}^{(N)}\right\rangle\right| + \max_{1\le k\le N-|\mathcal{Y}_L|-\left|\Lambda_{m-1}^f\right|}\left|\left\langle \mathbf{x}_{f_k}, \mathbf{r}_{m-1,\perp}^{(N)}\right\rangle\right| + \max_{1\le k\le N-|\mathcal{Y}_L|-\left|\Lambda_{m-1}^f\right|}\left|\left\langle \mathbf{e}_{f_k}, \mathbf{r}_{m-1}^{(N)}\right\rangle\right| \\ &\overset{(a)}{\le} \left|\left\langle \mathbf{x}_{\bar{f}_{p-k_m+1}}, \mathbf{r}_{m-1,\parallel}^{(N)}\right\rangle\right| + \max_{1\le k\le N-|\mathcal{Y}_L|-\left|\Lambda_{m-1}^f\right|}\left|\left\langle \mathbf{x}_{f_k}, \mathbf{r}_{m-1,\perp}^{(N)}\right\rangle\right| + \max_{1\le k\le N-|\mathcal{Y}_L|-\left|\Lambda_{m-1}^f\right|}\left|\left\langle \mathbf{e}_{f_k}, \mathbf{r}_{m-1}^{(N)}\right\rangle\right|,\end{aligned} \tag{5.10}$$

where (a) is true due to the ordering of $\left|\left\langle \mathbf{x}_{\bar{f}_h}, \mathbf{r}_{m-1,\parallel}^{(N)}\right\rangle\right|$'s. Therefore, at most $p-k_m$ absolute inner products $\left|\left\langle \mathbf{y}_{\bar{f}_u}, \mathbf{r}_{m-1}^{(N)}\right\rangle\right|$ are no smaller than the RHS of (5.10). As a result, for $\left|\left\langle \mathbf{y}_{f_{p-k_m+1}}, \mathbf{r}_{m-1}^{(N)}\right\rangle\right|$, the $p-k_m+1$ th largest $\left|\left\langle \mathbf{y}_{f_k}, \mathbf{r}_{m-1}^{(N)}\right\rangle\right|$, we must have

$$\left|\left\langle \mathbf{y}_{f_{p-k_m+1}}, \mathbf{r}_{m-1}^{(N)}\right\rangle\right| \le \left|\left\langle \mathbf{x}_{\bar{f}_{p-k_m+1}}, \mathbf{r}_{m-1,\parallel}^{(N)}\right\rangle\right| + \max_{1\le k\le N-|\mathcal{Y}_L|-\left|\Lambda_{m-1}^f\right|}\left|\left\langle \mathbf{x}_{f_k}, \mathbf{r}_{m-1,\perp}^{(N)}\right\rangle\right| + \max_{1\le k\le N-|\mathcal{Y}_L|-\left|\Lambda_{m-1}^f\right|}\left|\left\langle \mathbf{e}_{f_k}, \mathbf{r}_{m-1}^{(N)}\right\rangle\right|. \tag{5.11}$$

Now we have a lower bound for $\left|\left\langle \mathbf{y}_{t_{k_m}}, \mathbf{r}_{m-1}^{(N)}\right\rangle\right|$ in (5.9) and an upper bound for $\left|\left\langle \mathbf{y}_{f_{p-k_m+1}}, \mathbf{r}_{m-1}^{(N)}\right\rangle\right|$ in (5.11). Putting them together, the condition $\left|\left\langle \mathbf{y}_{f_{p-k_m+1}}, \mathbf{r}_{m-1}^{(N)}\right\rangle\right| \le \left|\left\langle \mathbf{y}_{t_{k_m}}, \mathbf{r}_{m-1}^{(N)}\right\rangle\right|$ is guaranteed, and hence the event $E_m$ occurs, once the following inequality is true

$$\begin{aligned}&\left|\left\langle \mathbf{x}_{\bar{f}_{p-k_m+1}}, \mathbf{r}_{m-1,\parallel}^{(N)}\right\rangle\right| + \max_{1\le k\le N-|\mathcal{Y}_L|-\left|\Lambda_{m-1}^f\right|}\left|\left\langle \mathbf{x}_{f_k}, \mathbf{r}_{m-1,\perp}^{(N)}\right\rangle\right| + \max_{1\le k\le N-|\mathcal{Y}_L|-\left|\Lambda_{m-1}^f\right|}\left|\left\langle \mathbf{e}_{f_k}, \mathbf{r}_{m-1}^{(N)}\right\rangle\right| \\ &\qquad + \max_{1\le j\le |\mathcal{Y}_L|-1-\left|\Lambda_{m-1}^t\right|}\left|\left\langle \mathbf{e}_{t_j}, \mathbf{r}_{m-1}^{(N)}\right\rangle\right| \le \left|\left\langle \mathbf{x}_{\bar{t}_{k_m}}, \mathbf{r}_{m-1,\parallel}^{(N)}\right\rangle\right|.\end{aligned} \tag{5.12}$$

Recall that our goal is to show that $\bigcap_{m=1}^{M} E_m$ occurs with a high probability. Under the semi-random model, we go on to estimate the probability of which the inequality (5.12) holds for all $1 \le m \le M$, and in turn a lower bound for $\Pr\{\bigcap_{m=1}^{M} E_m\}$. The basic idea is to "split and then lump": find an upper bound for each individual left-hand-side (LHS) term of (5.12) and estimate, one by one, the probability about which the obtained inequality holds, and similarly a lower bound for the RHS term along with an estimated probability of its validity; putting the results altogether gives a sufficient condition for (5.12) along with the desired probability lower bound.

*(i) An Upper Bound for the 1st LHS Term of (5.12):* For a fixed $\alpha$, we first note that

$$\begin{aligned}&\Pr\left\{\left|\left\langle \mathbf{x}_{\bar{f}_{p-k_m+1}}, \mathbf{r}_{m-1,\parallel}^{(N)}\right\rangle\right| \le \alpha\right\} \\ &= 1-\Pr\left\{\left|\left\langle \mathbf{x}_{\bar{f}_{p-k_m+1}}, \mathbf{r}_{m-1,\parallel}^{(N)}\right\rangle\right| > \alpha\right\} \\ &\overset{(a)}{\ge} 1-\Pr\left\{\exists I \subset \left[N-|\mathcal{Y}_L|-\left|\Lambda_{m-1}^f\right|\right] \text{ with } |I| = p-k_m+1 \text{ s.t. } \left|\left\langle \mathbf{x}_{f_k}, \mathbf{r}_{m-1,\parallel}^{(N)}\right\rangle\right| > \alpha, \forall k\in I\right\} \\ &\overset{(b)}{\ge} 1-\binom{N-|\mathcal{Y}_L|}{p-k_m+1}\left(\max_{\substack{I:|I|=p-k_m+1\\ I\subset\left[N-|\mathcal{Y}_L|-\left|\Lambda_{m-1}^f\right|\right]}} \Pr\left\{\left|\left\langle \mathbf{x}_{f_k}, \mathbf{r}_{m-1,\parallel}^{(N)}\right\rangle\right| > \alpha, \forall k\in I\right\}\right) \\ &\overset{(c)}{\ge} 1-\left(\frac{e(N-|\mathcal{Y}_L|)}{p-k_m+1}\right)^{(p-k_m+1)}\left(\max_{\substack{I:|I|=p-k_m+1\\ I\subset\left[N-|\mathcal{Y}_L|-\left|\Lambda_{m-1}^f\right|\right]}} \Pr\left\{\left|\left\langle \mathbf{x}_{f_k}, \mathbf{r}_{m-1,\parallel}^{(N)}\right\rangle\right| > \alpha, \forall k\in I\right\}\right),\end{aligned} \tag{5.13}$$

where in (a) the notation $[Q] \triangleq \{1,2,\cdots,Q\}$, $Q \in \mathbb{N}$, in (b) $\binom{Q}{k} \triangleq \frac{Q!}{(Q-k)!\,k!}$, and (c) holds by [25]. By Lemma 5.2 and since data points are independent, for $|I| = p - k_m + 1$ we have

$$\Pr\left\{\left|\left\langle \mathbf{x}_{f_k}, \mathbf{r}_{m-1,\|}^{(N)} \right\rangle\right| > \max_{l:l\neq L} 4\log(N) \frac{\left\|\mathbf{U}_l^T \mathbf{U}_L\right\|_F}{\sqrt{d_l d_L}} \left\|\mathbf{r}_{m-1,\|}^{(N)}\right\|_2, \forall k \in I\right\} \leq \left(\frac{2}{N^{8\log N/d_L}}\right)^{(p-k_m+1)}, \tag{5.14}$$

Combining (5.13) and (5.14) and setting $\alpha = \max_{l:l\neq L} 4\log(N)\left\|\mathbf{U}_l^T\mathbf{U}_L\right\|_F \left\|\mathbf{r}_{m-1,\|}^{(N)}\right\|_2 / \sqrt{d_l d_L}$, it then follows

$$\left|\left\langle \mathbf{x}_{\bar{f}_{p-k_m+1}}, \mathbf{r}_{m-1,\|}^{(N)} \right\rangle\right| \leq \max_{l:l\neq L} 4\log(N) \frac{\left\|\mathbf{U}_l^T \mathbf{U}_L\right\|_F}{\sqrt{d_l d_L}} \left\|\mathbf{r}_{m-1,\|}^{(N)}\right\|_2 \tag{5.15}$$

holds with a probability at least $1 - \left(\frac{2e(N-|\mathcal{Y}_L|)}{(p-k_m+1)\,N^{8\log N/d_L}}\right)^{(p-k_m+1)}$

<u>*(ii) An Upper Bound for the 2nd LHS Term of (5.12):*</u> We first note that

$$\begin{aligned}
&\Pr\left\{\max_{1\leq k\leq N-|\mathcal{Y}_L|-\left|\Lambda_{m-1}^f\right|} \left|\left\langle \mathbf{x}_{f_k}, \mathbf{r}_{m-1,\perp}^{(N)}\right\rangle\right| \leq \alpha\right\} \\
&= 1 - \Pr\left\{\max_{1\leq k\leq N-|\mathcal{Y}_L|-\left|\Lambda_{m-1}^f\right|} \left|\left\langle \mathbf{x}_{f_k}, \mathbf{r}_{m-1,\perp}^{(N)}\right\rangle\right| > \alpha\right\} \\
&\geq 1 - \Pr\left\{\exists k \in \left[N - |\mathcal{Y}_L| - \left|\Lambda_{m-1}^f\right|\right] \text{ s.t. } \left|\left\langle \mathbf{x}_{f_k}, \mathbf{r}_{m-1,\perp}^{(N)}\right\rangle\right| > \alpha\right\} \\
&= 1 - \Pr\left\{\left(\bigcap_{k=1}^{N-|\mathcal{Y}_L|-\left|\Lambda_{m-1}^f\right|} \left\{\left|\left\langle \mathbf{x}_{f_k}, \mathbf{r}_{m-1,\perp}^{(N)}\right\rangle\right| \leq \alpha\right\}\right)^c\right\} \\
&= 1 - \Pr\left\{\bigcup_{k=1}^{N-|\mathcal{Y}_L|-\left|\Lambda_{m-1}^f\right|} \left\{\left|\left\langle \mathbf{x}_{f_k}, \mathbf{r}_{m-1,\perp}^{(N)}\right\rangle\right| > \alpha\right\}\right\} \\
&\geq 1 - \sum_{k=1}^{N-|\mathcal{Y}_L|-\left|\Lambda_{m-1}^f\right|} \Pr\left\{\left|\left\langle \mathbf{x}_{f_k}, \mathbf{r}_{m-1,\perp}^{(N)}\right\rangle\right| > \alpha\right\}.
\end{aligned} \tag{5.16}$$

By Lemma 5.3, we have

$$\Pr\left\{\left|\left\langle \mathbf{x}_{f_k}, \mathbf{r}_{m-1,\perp}^{(N)}\right\rangle\right| > \sqrt{\frac{6\log N}{n-d_L}} \left\|\mathbf{r}_{m-1,\perp}^{(N)}\right\|_2\right\} \leq \frac{2c}{N^3},\ \ 1 \leq k \leq N - |\mathcal{Y}_L| - \left|\Lambda_{m-1}^f\right|, \tag{5.17}$$

Combing (5.16) and (5.17) and with $\alpha = \sqrt{\frac{6\log N}{n-d_L}} \left\|\mathbf{r}_{m-1,\perp}^{(N)}\right\|_2$, we obtain

$$\max_{1\leq k\leq N-|\mathcal{Y}_L|-\left|\Lambda_{m-1}^f\right|} \left|\left\langle \mathbf{x}_{f_k}, \mathbf{r}_{m-1,\perp}^{(N)}\right\rangle\right| \leq \sqrt{\frac{6\log N}{n-d_L}} \left\|\mathbf{r}_{m-1,\perp}^{(N)}\right\|_2 \tag{5.18}$$

holds with a probability at least $1 - \frac{2c}{N^2}$.

*(iii) An Upper Bound for the 3rd LHS Term of (5.12):* Using similar techniques as in deriving (5.16) we can first reach

$$\begin{aligned}&\Pr\left\{\max_{1\le k\le N-|\mathcal{Y}_L|-|\Lambda_{m-1}^f|}\left|\left\langle \mathbf{e}_{f_k},\mathbf{r}_{m-1}^{(N)}\right\rangle\right|\le \alpha\right\}\\&\ge 1-\sum_{k=1}^{N-|\mathcal{Y}_L|-|\Lambda_{m-1}^f|}\Pr\left\{\left|\left\langle \mathbf{e}_{f_k},\mathbf{r}_{m-1}^{(N)}\right\rangle\right|>\alpha\right\}.\end{aligned} \tag{5.19}$$

Using part (a) of Lemma 5.1 and with $\mathbf{a}=\mathbf{e}_{f_k}/\|\mathbf{e}_{f_k}\|_2$ and $\mathbf{b}=\mathbf{r}_{m-1}^{(N)}$, it follows

$$\begin{aligned}&\Pr\left\{\left|\left\langle \mathbf{e}_{f_k},\mathbf{r}_{m-1}^{(N)}\right\rangle\right|>\sqrt{\frac{6\log N}{n}}\|\mathbf{r}_{m-1}^{(N)}\|_2\|\mathbf{e}_{f_k}\|_2\right\}\\&=\Pr\left\{\left|\left\langle \frac{\mathbf{e}_{f_k}}{\|\mathbf{e}_{f_k}\|_2},\mathbf{r}_{m-1}^{(N)}\right\rangle\right|>\sqrt{\frac{6\log N}{n}}\|\mathbf{r}_{m-1}^{(N)}\|_2\right\}\le\frac{2}{N^3}.\end{aligned} \tag{5.20}$$

Under Assumption 1 we have

$$\|\mathbf{r}_{m-1}^{(N)}\|_2=\left\|\mathrm{P}_{\mathcal{R}(\mathbf{Y}_{\Lambda_{m-1}})}\mathbf{y}_N\right\|_2\le\|\mathbf{y}_N\|_2=\|\mathbf{x}_N+\mathbf{e}_N\|_2\le 1+\|\mathbf{e}_N\|_2, \tag{5.21}$$

in which $\mathcal{R}(\mathbf{Y}_{\Lambda_{m-1}})$ is the column space of $\mathbf{Y}_{\Lambda_{m-1}}$ and the last inequality holds by the triangle inequality. Combing (5.20) and (5.21) yields

$$\Pr\left\{\left|\left\langle \mathbf{e}_{f_k},\mathbf{r}_{m-1}^{(N)}\right\rangle\right|>\sqrt{\frac{6\log N}{n}}(1+\|\mathbf{e}_N\|_2)\max_{1\le k\le N-|\mathcal{Y}_L|-|\Lambda_{m-1}^f|}\|\mathbf{e}_{f_k}\|_2\right\}\le\frac{2}{N^3}. \tag{5.22}$$

Setting $\alpha=\sqrt{\frac{6\log N}{n}}(1+\|\mathbf{e}_N\|_2)\max_{1\le k\le N-|\mathcal{Y}_L|-|\Lambda_{m-1}^f|}\|\mathbf{e}_{f_k}\|_2$, (5.19) and (5.22) imply

$$\max_{1\le k\le N-|\mathcal{Y}_L|-|\Lambda_{m-1}^f|}\left|\left\langle \mathbf{e}_{f_k},\mathbf{r}_{m-1}^{(N)}\right\rangle\right|\le\sqrt{\frac{6\log N}{n}}(1+\|\mathbf{e}_N\|_2)\max_{1\le k\le N-|\mathcal{Y}_L|-|\Lambda_{m-1}^f|}\|\mathbf{e}_{f_k}\|_2 \tag{5.23}$$

holds with a probability at least $1-\frac{2}{N^2}$.

*(iv) An Upper Bound for the 4th LHS Term of (5.12):* By following the same procedures as from (5.19) to (5.23), we can readily show the inequality

$$\max_{1\le j\le|\mathcal{Y}_L|-1-|\Lambda_{m-1}^t|}\left|\left\langle \mathbf{e}_{t_j},\mathbf{r}_{m-1}^{(N)}\right\rangle\right|\le\sqrt{\frac{6\log N}{n}}(1+\|\mathbf{e}_N\|_2)\max_{1\le j\le|\mathcal{Y}_L|-1-|\Lambda_{m-1}^t|}\|\mathbf{e}_{t_j}\|_2, \tag{5.24}$$

holds with a probability at least $1-\frac{2}{N^2}$.

*(v) A lower Bound for the RHS Term of (5.12):* Next, we go on to find a lower bound for the RHS term of (5.12). Using part (b) of Lemma 5.1 and with $\mathbf{a}=\mathbf{x}_{t_j}$ and $\mathbf{b}=\mathbf{r}_{m-1,\|}^{(N)}$, we have

$$\Pr\left\{\left|\left\langle \mathbf{x}_{t_j},\mathbf{r}_{m-1,\|}^{(N)}\right\rangle\right|<\frac{\tau}{\sqrt{d_L}}\left\|\mathbf{r}_{m-1,\|}^{(N)}\right\|_2\right\}\le\sqrt{\frac{2}{\pi}}\tau,\ \ 1\le j\le|\mathcal{Y}_L|-1-|\Lambda_{m-1}^t|, \tag{5.25}$$

which together with the assumption that $\mathbf{x}_{t_j}$ 's are independent yields

$$\Pr\left\{\left|\left\langle \mathbf{x}_{t_j}, \mathbf{r}_{m-1,\|}^{(N)}\right\rangle\right| < \frac{\tau}{\sqrt{d_L}}\left\|\mathbf{r}_{m-1,\|}^{(N)}\right\|_2, \forall j \in I\right\} \le \left(\sqrt{\frac{2}{\pi}}\tau\right)^{(|\mathcal{Y}_L|-|\Lambda_{m-1}^t|-k_m)}, \quad I \subset \left[|\mathcal{Y}_L|-1-\left|\Lambda_{m-1}^t\right|\right], \ |I| = |\mathcal{Y}_L| - \left|\Lambda_{m-1}^t\right| - k_m . \tag{5.26}$$

It then follows

$$\begin{aligned}
&\Pr\left\{\left|\left\langle \mathbf{x}_{t_{\bar{k}_m}}, \mathbf{r}_{m-1,\|}^{(N)}\right\rangle\right| \ge \frac{\tau}{\sqrt{d_L}}\left\|\mathbf{r}_{m-1,\|}^{(N)}\right\|_2\right\} \\
&= 1 - \Pr\left\{\left|\left\langle \mathbf{x}_{t_{\bar{k}_m}}, \mathbf{r}_{m-1,\|}^{(N)}\right\rangle\right| < \frac{\tau}{\sqrt{d_L}}\left\|\mathbf{r}_{m-1,\|}^{(N)}\right\|_2\right\} \\
&= 1 - \Pr\left\{\exists I \subset \left[|\mathcal{Y}_L|-1-\left|\Lambda_{m-1}^t\right|\right] \text{ with } |I| = |\mathcal{Y}_L|-1-\left|\Lambda_{m-1}^t\right|-k_m+1 \text{ s.t. } \left|\left\langle \mathbf{x}_{t_j}, \mathbf{r}_{m-1,\|}^{(N)}\right\rangle\right| < \frac{\tau}{\sqrt{d_L}}\left\|\mathbf{r}_{m-1,\|}^{(N)}\right\|_2, \forall j \in I\right\} \\
&\ge 1 - \binom{|\mathcal{Y}_L|-1}{k_m-1} \max_{\substack{I:|I|=|\mathcal{Y}_L|-|\Lambda_{m-1}^t|-k_m \\ I\subset[|\mathcal{Y}_L|-1-|\Lambda_{m-1}^t|]}} \Pr\left\{\left|\left\langle \mathbf{x}_{t_j}, \mathbf{r}_{m-1,\|}^{(N)}\right\rangle\right| < \frac{\tau}{\sqrt{d_L}}\left\|\mathbf{r}_{m-1,\|}^{(N)}\right\|_2, \forall j \in I\right\} \\
&\overset{(a)}{\ge} 1 - \binom{|\mathcal{Y}_L|-1}{k_m-1} \max_{\substack{I:|I|=|\mathcal{Y}_L|-|\Lambda_{m-1}^t|-k_m \\ I\subset[|\mathcal{Y}_L|-1-|\Lambda_{m-1}^t|]}} \left(\sqrt{\frac{2}{\pi}}\tau\right)^{(|\mathcal{Y}_L|-|\Lambda_{m-1}^t|-k_m)} \\
&\ge 1 - \left(\frac{e(|\mathcal{Y}_L|-1)}{k_m-1}\right)^{k_m-1}\left(\sqrt{\frac{2}{\pi}}\tau\right)^{(|\mathcal{Y}_L|-d_L-k_m)},
\end{aligned} \tag{5.27}$$

where (a) holds thanks to (5.26). Hence, the inequality

$$\left|\left\langle \mathbf{x}_{t_{\bar{k}_m}}, \mathbf{r}_{m-1,\|}^{(N)}\right\rangle\right| \ge \tau\left\|\mathbf{r}_{m-1,\|}^{(N)}\right\|_2 / \sqrt{d_L} \tag{5.28}$$

holds with a probability at least $1-\left(\frac{e(|\mathcal{Y}_L|-1)}{k_m-1}\right)^{k_m-1}\left(\sqrt{\frac{2}{\pi}}\tau\right)^{(|\mathcal{Y}_L|-d_L-k_m)}$.

With the bounds in (5.15), (5.18), (5.23), (5.24) and (5.28), the inequality in (5.12) is true, and hence the event $E_m$ occurs, when the following inequality holds:

$$\begin{aligned}
&\max_{l:l\ne L} 4\log(N)\frac{\left\|\mathbf{U}_l^T\mathbf{U}_L\right\|_F}{\sqrt{d_l d_L}}\left\|\mathbf{r}_{m-1,\|}^{(N)}\right\|_2 + \sqrt{\frac{6\log N}{n-d_L}}\left\|\mathbf{r}_{m-1,\perp}^{(N)}\right\|_2 \\
&+\sqrt{\frac{6\log N}{n}}(1+\|\mathbf{e}_N\|_2)\left(\max_{1\le k\le N-|\mathcal{Y}_L|-|\Lambda_{m-1}^f|}\left\|\mathbf{e}_{f_k}\right\|_2 + \max_{1\le j\le|\mathcal{Y}_L|-1-|\Lambda_{m-1}^t|}\left\|\mathbf{e}_{t_j}\right\|_2\right) \le \frac{\tau}{\sqrt{d_L}}\left\|\mathbf{r}_{m-1,\|}^{(N)}\right\|_2,
\end{aligned} \tag{5.29}$$

or equivalently,

$$\begin{aligned}
&\max_{l:l\ne L}\frac{\left\|\mathbf{U}_l^T\mathbf{U}_L\right\|_F}{\sqrt{d_l}} + \frac{\sqrt{3d_L}\left\|\mathbf{r}_{m-1,\perp}^{(N)}\right\|_2}{\sqrt{8(n-d_L)\log N}\left\|\mathbf{r}_{m-1,\|}^{(N)}\right\|_2} \\
&+\frac{\sqrt{3d_L}}{\sqrt{8n\log N}\left\|\mathbf{r}_{m-1,\|}^{(N)}\right\|_2}(1+\|\mathbf{e}_N\|_2)\left(\max_{1\le k\le N-|\mathcal{Y}_L|-|\Lambda_{m-1}^f|}\left\|\mathbf{e}_{f_k}\right\|_2 + \max_{1\le j\le|\mathcal{Y}_L|-1-|\Lambda_{m-1}^t|}\left\|\mathbf{e}_{t_j}\right\|_2\right) \le \frac{\tau}{4\log N},
\end{aligned} \tag{5.30}$$

which is obtained by multiplying the inequality (5.29) throughout by the factor $\sqrt{d_L}/\left(4\log N\left\|\mathbf{r}_{m-1,\|}^{(N)}\right\|_2\right)$.

It can then be concluded that $\bigcap_{m=1}^{M} E_m$ occurs once (5.15), (5.18), (5.23), (5.24) (5.28) and (5.30) hold for all $1 \le m \le M$. Hence, a lower bound for $\Pr\left\{\bigcap_{m=1}^{M} E_m\right\}$ can be obtained by finding a lower bound for the probability that (5.30) holds for all $1 \le m \le M$.

Still, our approach seeks an upper bound for the LHS term in (5.30), hence a sufficient condition guaranteeing (5.30), and proves this bound holds with a high probability. Towards this end, we derive an upper bound for $\left\|\mathbf{r}_{m-1,\perp}^{(N)}\right\|_2$, $1 \le m \le M$, an upper bound for $\|\mathbf{e}_i\|_2$ $1 \le i \le N$, and a lower bound for $\left\|\mathbf{r}_{m-1,\|}^{(N)}\right\|_2$, $1 \le m \le M$; lumping these altogether yields the claimed result. To begin with, we write

$$\left\|\mathbf{r}_{m-1,\perp}^{(N)}\right\|_2 = \left\|\mathrm{P}_{\mathcal{S}_L^\perp}\left(\mathbf{I} - \mathbf{Y}_{\Lambda_{m-1}}\mathbf{Y}_{\Lambda_{m-1}}^\dagger\right)\mathbf{y}_N\right\|_2 \le \left\|\mathrm{P}_{\mathcal{S}_L^\perp}\mathbf{y}_N\right\|_2 \le \|\mathbf{e}_N\|_2 . \tag{5.31}$$

According to [20, Lemma 10], the event

$$\bigcap_{i=1}^{N}\{\|\mathbf{e}_i\|_2 \le 3\sigma/2\} \tag{5.32}$$

occurs with a probability at least $1 - Ne^{-n/8}$. Hence, we conclude that the following sets of inequalities

$$\left\|\mathbf{r}_{m-1,\perp}^{(N)}\right\|_2 \le 3\sigma/2,\ 1 \le m \le M, \tag{5.33}$$

$$\|\mathbf{e}_i\|_2 \le 3\sigma/2,\ 1 \le i \le N, \tag{5.34}$$

hold at once with a probability at least $1 - Ne^{-n/8}$. Next, we will find a lower bound for $\left\|\mathbf{r}_{m-1,\|}^{(N)}\right\|_2$'s. For $1 \le m \le M$, we have

$$\begin{aligned}\mathbf{r}_{m-1,\|}^{(N)} &= \mathrm{P}_{\mathcal{S}_L}\left((\mathbf{I} - \mathbf{Y}_{\Lambda_{m-1}}(\mathbf{Y}_{\Lambda_{m-1}})^\dagger)\mathbf{y}_N\right)\\ &=\mathrm{P}_{\mathcal{S}_L}\left(\mathbf{y}_N - \mathbf{Y}_{\Lambda_{m-1}}\mathbf{c}_{m-1}^*\right)\\ &=\mathbf{y}_{N,\|} - \mathbf{Y}_{\Lambda_{m-1,\|}}\mathbf{c}_{m-1}^*,\end{aligned} \tag{5.35}$$

where $\mathbf{y}_{N,\|} \triangleq \mathrm{P}_{\mathcal{S}_L}\mathbf{y}_N$, $\mathbf{Y}_{\Lambda_{m-1,\|}} \triangleq \mathrm{P}_{\mathcal{S}_L}\mathbf{Y}_{\Lambda_{m-1}}$ and $\mathbf{c}_{m-1}^* \triangleq \arg\min_{\mathbf{c}_{m-1}}\left\|\mathbf{y}_N - \mathbf{Y}_{\Lambda_{m-1}}\mathbf{c}_{m-1}\right\|_2$. Then a lower bound for $\left\|\mathbf{r}_{m-1,\|}^{(N)}\right\|_2$ is obtained as

$$\begin{aligned}\left\|\mathbf{r}_{m-1,\|}^{(N)}\right\|_2 &= \left\|\mathbf{y}_{N,\|} - \mathbf{Y}_{\Lambda_{m-1,\|}}\mathbf{c}_{m-1}^*\right\|_2\\ &\ge \min_{\mathbf{c}}\left\|\mathbf{y}_{N,\|} - \mathbf{Y}_{\Lambda_{m-1,\|}}\mathbf{c}\right\|_2\\ &=\left\|\mathbf{y}_{N,\|} - \mathbf{Y}_{\Lambda_{m-1,\|}}\mathbf{Y}_{\Lambda_{m-1,\|}}^\dagger\mathbf{y}_{N,\|}\right\|_2\\ &\ge\left\|\mathbf{y}_{N,\|} - \mathbf{Y}_{\Lambda_{M-1,\|}}\mathbf{Y}_{\Lambda_{M-1,\|}}^\dagger\mathbf{y}_{N,\|}\right\|_2\\ &=\left\|\mathrm{P}_{\mathcal{B}}\mathbf{y}_{N,\|}\right\|_2,\end{aligned} \tag{5.36}$$

where $\mathcal{B}= \mathcal{S}_L \cap \mathcal{R}(\mathbf{Y}_{\Lambda_{M-1,\|}})^\perp$ is a subspace of dimension $d_L - p(M-1) > 0$. Since $\mathbf{y}_{N,\|} = \mathbf{x}_N + \mathbf{e}_{N,\|}$ and both the distributions of $\mathbf{x}_N$ and $\mathbf{e}_{N,\|}$ are rotationally invariant in $\mathcal{S}_L$ [20, p.4092], the normalized $\mathbf{y}_{N,\|}/\left\|\mathbf{y}_{N,\|}\right\|_2$ is uniformly distributed over $\mathcal{S}_L \cap \mathbb{S}^{n-1}$. Let $\mathbf{V}_{\mathcal{B}} = [\mathbf{v}_1\ \mathbf{v}_2 \cdots \mathbf{v}_{d_L - p(M-1)}] \in \mathbb{R}^{n\times(d_L - p(M-1))}$ be a matrix whose columns form an orthonormal basis for $\mathcal{B}$; augment $\mathbf{V}_{\mathcal{B}}$ by adding extra $p(M-1)$ columns so that the $d_L$ columns of $\mathbf{V}=[\mathbf{v}_1\ \mathbf{v}_2\cdots\mathbf{v}_{d_L-p(M-1)}\cdots\mathbf{v}_{d_L}]\in\mathbb{R}^{n\times d_L}$ form an orthonormal basis for $\mathcal{S}_L$. Clearly, we have $\mathbf{y}_{N,\|}/\left\|\mathbf{y}_{N,\|}\right\|_2=\mathbf{V}\mathbf{a}$, where $\mathbf{a}=[a_1\ a_2\cdots a_{d_L}]^T$ obeys uniform distribution over the unit sphere of $\mathbb{R}^{d_L}$ and, in particular, $\mathrm{P}_{\mathcal{B}}(\mathbf{y}_{N,\|}/\left\|\mathbf{y}_{N,\|}\right\|_2)=\mathbf{V}_{\mathcal{B}}\tilde{\mathbf{a}}$, where $\tilde{\mathbf{a}}=[a_1\ a_2\cdots a_{d_L-p(M-1)}]^T$ obeys the distribution specified in [26, eq. (7)]. Noticing $\left\|\mathrm{P}_{\mathcal{B}}\mathbf{y}_{N,\|}/\left\|\mathbf{y}_{N,\|}\right\|_2\right\|_2=\|\mathbf{V}_{\mathcal{B}}\tilde{\mathbf{a}}\|_2=\|\tilde{\mathbf{a}}\|_2$, the following set of inequalities hold

$$
\begin{aligned}
\Pr\left\{\left\|\mathrm{P}_{\mathcal{B}}\frac{\mathbf{y}_{N,\|}}{\left\|\mathbf{y}_{N,\|}\right\|_2}\right\|_2\le\lambda\right\}&=\Pr\{\|\tilde{\mathbf{a}}\|_2\le\lambda\}\\
&\overset{(a)}{=}\int_{\|\tilde{\mathbf{a}}\|_2\le\lambda}\frac{\Gamma(d_L/2)}{\pi^{(d_L-p(M-1))/2}\Gamma(p(M-1)/2)}(1-\|\tilde{\mathbf{a}}\|_2^2)^{(p(M-1)-2)/2}d\tilde{\mathbf{a}}\\
&=\frac{\Gamma(d_L/2)}{\pi^{(d_L-p(M-1))/2}\Gamma(p(M-1)/2)}\int_{\|\tilde{\mathbf{a}}\|_2\le\lambda}(1-\|\tilde{\mathbf{a}}\|_2^2)^{(p(M-1)-2)/2}d\tilde{\mathbf{a}}\\
&\le\frac{\Gamma(d_L/2)}{\pi^{(d_L-p(M-1))/2}\Gamma(p(M-1)/2)}v(d_L-p(M-1))\lambda^{d_L-p(M-1)}\\
&\overset{(b)}{\le}\left(\frac{d_L}{2\pi}\right)^{(d_L-p(M-1))/2}v(d_L-p(M-1))\lambda^{d_L-p(M-1)},
\end{aligned}
\tag{5.37}
$$

where (a) holds by [26, eq. (7)] and (b) follows from [27, eq. (8.1)]. Using (5.36) and (5.37) with $\lambda=\sigma\sqrt{2}/\sqrt{d_L}$, we can obtain

$$
\left\|\mathbf{r}_{m-1,\|}^{(N)}\right\|_2>\left(\sigma\sqrt{2}\,\|\mathbf{y}_{N,\|}\|_2\right)/\sqrt{d_L},\ \ 1\le m\le M,
\tag{5.38}
$$

holds with a probability at least $1-v(d_L-p(M-1))(\sigma/\sqrt{\pi})^{d_L-p(M-1)}$. Since $\|\mathbf{y}_{N,\|}\|_2=\|\mathbf{x}_N+\mathbf{e}_{N,\|}\|_2$ and $\|\mathbf{x}_N\|_2=1$ (see Assumption 1), triangle inequality gives

$$
\|\mathbf{y}_{N,\|}\|_2\ge1-\left\|\mathbf{e}_{N,\|}\right\|_2.
\tag{5.39}
$$

Combining (5.38) and (5.39), we have

$$
\left\|\mathbf{r}_{m-1,\|}^{(N)}\right\|_2>\left[\sigma\sqrt{2}(1-\left\|\mathbf{e}_{N,\|}\right\|_2)\right]/\sqrt{d_L},\ \ 1\le m\le M,
\tag{5.40}
$$

holds with a probability as high as (5.38). With the aid of (5.33), (5.34), and (5.40), we go on to find a sufficient condition ensuring (5.30) by finding an upper bound for the LHS of (5.30). Once (5.33), (5.34) and (5.40) hold, we can obtain the following set of inequalities:

$$
\begin{aligned}
&\max_{l:l\ne L}\frac{\left\|\mathbf{U}_l^T\mathbf{U}_L\right\|_F}{\sqrt{d_l}}+\frac{\sqrt{3d_L}\left\|\mathbf{r}_{m-1,\perp}^{(N)}\right\|_2}{\sqrt{8(n-d_L)\log N}\left\|\mathbf{r}_{m-1,\|}^{(N)}\right\|_2}\\
&\quad+\frac{\sqrt{3d_L}}{\sqrt{8n\log N}\left\|\mathbf{r}_{m-1,\|}^{(N)}\right\|_2}(1+\|\mathbf{e}_N\|_2)\left(\max_{1\le k\le N-|\mathcal{Y}_L|-\left|\Lambda_{m-1}^f\right|}\left\|\mathbf{e}_{f_k}\right\|_2+\max_{1\le j\le|\mathcal{Y}_L|-1-\left|\Lambda_{m-1}^t\right|}\left\|\mathbf{e}_{t_j}\right\|_2\right)\\
&\overset{(a)}{\le}\max_{l:l\ne L}\frac{\left\|\mathbf{U}_l^T\mathbf{U}_L\right\|_F}{\sqrt{d_l}}+\frac{\sqrt{3d_L}\left\|\mathbf{r}_{m-1,\perp}^{(N)}\right\|_2}{\sqrt{8(n-d_L)\log N}\left\|\mathbf{r}_{m-1,\|}^{(N)}\right\|_2}+\frac{\sqrt{3d_L}}{\sqrt{8n\log N}\left\|\mathbf{r}_{m-1,\|}^{(N)}\right\|_2}(1+\frac{3\sigma}{2})3\sigma
\end{aligned}
$$

$$
\begin{aligned}
&\overset{(b)}{\leq} \max_{l:l\neq L} aff(\mathcal{S}_l,\mathcal{S}_L) + \frac{\sqrt{3d_L}\left\|\mathbf{r}_{m-1,\perp}^{(N)}\right\|_2}{\sqrt{8(n-d_L)\log N}\left\|\mathbf{r}_{m-1,\|}^{(N)}\right\|_2} + \frac{\sqrt{3d_L}}{\sqrt{8n\log N}\left\|\mathbf{r}_{m-1,\|}^{(N)}\right\|_2}(1+\frac{3\sigma}{2})3\sigma \\
&\overset{(c)}{\leq} \max_{l:l\neq L} aff(\mathcal{S}_l,\mathcal{S}_L) + \frac{\sqrt{3d_L}(3\sigma/2)}{\sqrt{8(n-d_L)\log N}\left\|\mathbf{r}_{m-1,\|}^{(N)}\right\|_2} + \frac{\sqrt{3d_L}}{\sqrt{8n\log N}\left\|\mathbf{r}_{m-1,\|}^{(N)}\right\|_2}(1+\frac{3\sigma}{2})3\sigma \\
&\overset{(d)}{\leq} \max_{l:l\neq L} aff(\mathcal{S}_l,\mathcal{S}_L) + \frac{\sqrt{3}d_L(3\sigma/2)}{\sqrt{8(n-d_L)\log N}\sigma\sqrt{2}(1-3\sigma/2)} + \frac{\sqrt{3}d_L}{\sqrt{8n\log N}\sigma\sqrt{2}(1-3\sigma/2)}(1+\frac{3\sigma}{2})3\sigma \\
&= \max_{l:l\neq L} aff(\mathcal{S}_l,\mathcal{S}_L) + \frac{3\sqrt{3}d_L}{(8-12\sigma)\sqrt{(n-d_L)\log N}} + \frac{3\sqrt{3}d_L(2+3\sigma)}{(8-12\sigma)\sqrt{n\log N}} \\
&\leq \max_{l:l\neq L} aff(\mathcal{S}_l,\mathcal{S}_L) + \frac{3\sqrt{3}d_L(3+3\sigma)}{(8-12\sigma)\sqrt{(n-d_L)\log N}} \\
&\overset{(e)}{\leq} \frac{\tau}{4\log N},
\end{aligned}
\tag{5.41}
$$

where (a) holds by (5.34), (b) follows from the definition (3.2) of affinity, (c) is due to (5.33) and (d) is obtained by using (5.34) and (5.40), and (e) is true thanks to Assumption 3. As a result, when (5.33), (5.34) and (5.40) hold, the inequality in (5.41) for all $1 \leq m \leq M$ is guaranteed. By employing the union bound technique it can be concluded that the inequality (5.30) for all $1 \leq m \leq M$ holds with a probability at least $1 - v(d_L - p(M-1))(\sigma/\sqrt{\pi})^{d_L - p(M-1)} - Ne^{-n/8}$. Finally, by using (5.15), (5.18), (5.23), (5.24), (5.28) and (5.30), the proof is thus completed again by using the union bound. □

*B. Proof of Corollary 3.2*

The lower bound (3.6) is simply obtained as the probability of the event that the numbers $k_m$'s of recovered true neighbors throughout all $M$ iterations yield the maximal lower bound (3.4). That is to say, (3.6) is the maximum of (3.4) over all plausible $0 \leq k_m \leq p$, $1 \leq m \leq M$; more precisely, the minimum objective of the following optimization problem

$$
\begin{aligned}
&\min_{(k_1,\cdots k_M)} J_0 = \sum_{m=1}^{M} J(k_m) \\
&\text{s.t. } \sum_{m=1}^{M} k_m - k_t = 0 \\
&\qquad k_m \in \{0,1,2,\cdots,p\},\ \forall 1 \leq m \leq M,
\end{aligned}
\tag{5.42}
$$

where

$$
J(k_m) \triangleq \left(\left(\frac{2e(N-|\mathcal{Y}_L|)}{(p-k_m+1)N^{8\log N/d_L}}\right)^{(p-k_m+1)} + \left(\sqrt{\frac{2}{\pi}}\tau\right)^{(|\mathcal{Y}_L|-d_L-k_m)}\left(\frac{e(|\mathcal{Y}_L|-1)}{k_m-1}\right)^{k_m-1}\right)\mathbf{1}(k_m>0). \tag{5.43}
$$

Below, we will show that the two-level sequence

$$
(k_1,\cdots,k_M) = (\underbrace{q_t+1, q_t+1, \cdots q_t+1}_{r_t\text{-fold}}, \underbrace{q_t, q_t, \cdots q_t}_{(M-r_t)\text{-fold}}) \tag{5.44}
$$

solves (5.42), consequently leading to (3.6). We recall the following definition and lemma, which are needed in our proof.

***Definition 5.4 [28]:*** Let $\mathbf{x} = [x_1 \ x_2 \cdots x_M]^T \in \mathbb{R}^M$ and $\mathbf{y} = [y_1 \ y_2 \cdots y_M]^T \in \mathbb{R}^M$ be two real vectors whose entries are ordered in the way that[1] $x_{[1]} \geq x_{[2]} \geq \cdots \geq x_{[M]}$ and $y_{[1]} \geq y_{[2]} \geq \cdots \geq y_{[M]}$, respectively. Then $\mathbf{x}$ is said to be majorized by $\mathbf{y}$ if

$$\sum_{m=1}^{s} x_{[m]} \leq \sum_{m=1}^{s} y_{[m]} \quad \text{for all} \quad 1 \leq s \leq M \tag{5.45}$$

and

$$\sum_{m=1}^{M} x_m = \sum_{m=1}^{M} y_m \,. \tag{5.46}$$

We say $f : \mathcal{A} \subset \mathbb{R}^M$ is Schur-convex if $f(\mathbf{x}) \leq f(\mathbf{y})$ whenever **x** is majorized by **y**, $\mathbf{x}, \mathbf{y} \in \mathcal{A}$. □

***Lemma 5.5 ([28, Theorem 12.25]):*** Let $f : \mathcal{A} \subset \mathbb{R}^M$ be a permutation invariant function, that is, $f(\mathbf{x}) = f(\mathbf{P}\mathbf{x})$ for all $\mathbf{x} \in \mathcal{A}$ and permutation matrices $\mathbf{P} \in \mathbb{R}^{M \times M}$, whose first partial derivatives exist in $\mathcal{A}$. Then $f$ is Schur-convex in $\mathcal{A}$ if and only if

$$(x_i - x_j)\left(\frac{\partial f}{\partial x_i} - \frac{\partial f}{\partial x_j}\right) \geq 0 \quad \text{for all} \quad \mathbf{x} = [x_1 \ x_2 \cdots x_M]^T \in \mathcal{A} \tag{5.47}$$

holds for all $1 \leq i \neq j \leq M$. □

Using vector-matrix notation, all we have to do is to show

$$\mathbf{k}^* \triangleq [\underbrace{q_t + 1 \ q_t + 1 \cdots q_t + 1}_{r_t\text{-fold}} \ \ \underbrace{q_t \ q_t \cdots q_t}_{(M - r_t)\text{-fold}}]^T \tag{5.48}$$

solves the following optimization problem:

$$\begin{aligned} &\min_{\mathbf{k} = [k_1, \cdots k_M]^T} \ J_0 = \sum_{m=1}^{M} J(\mathbf{k}^T \mathbf{v}_m) \\ &\text{s.t.} \ \ \mathbf{k}^T \mathbf{1} = k_t \\ &\qquad \mathbf{k}^T \mathbf{v}_m \in \{0, 1, 2, \cdots, p\}, \ \forall 1 \leq m \leq M, \end{aligned} \tag{5.49}$$

in which $\mathbf{v}_m \triangleq [\underbrace{0 \cdots 0}_{m-1 \text{ fold}} \ 1 \ 0 \cdots 0]^T$ is the $m$th standard unit vector. It suffices to prove

$$J_0\left(\mathbf{k}^*\right) \leq J_0\left(\mathbf{k}\right), \tag{5.50}$$

for any $\mathbf{k} \in \mathcal{D}_J$, the feasible set of (5.49), based on Lemma 5.5. For this we first note that the objective $J_0$ is not differentiable, since the function $J$ in (5.43) involves the indicator function. To rid of this difficulty, we consider the differentiable surrogate $\tilde{J} : \mathbb{R} \rightarrow \mathbb{R}$ for $J$, constructed according to

1. $x_{[i]}$ denotes the $i$th largest $x_j$, $1 \leq j \leq M$.

(i) $\tilde{J}(x) = J(x)$ for all $x \geq 0.9$ and $x = 0$. (5.51)

(ii) $\tilde{J}'(0) = J'(1)$. (5.52)

Thanks to (i), the function $\tilde{J}$ thus obtained satisfies $\tilde{J}(k_m) = J(k_m)$ for all $k_m \in \{0,1,2,\cdots,p\}$ and $\tilde{J}'(k_m) = J'(k_m)$ for all $k_m \in \{1,2,\cdots,p\}$. The corresponding differentiable surrogate for $J_0$ is accordingly given by

$$\tilde{J}_0(\mathbf{k}) = \sum_{m=1}^{M} \tilde{J}(\mathbf{k}^T \mathbf{v}_m). \tag{5.53}$$

Clearly, $\tilde{J}_0(\mathbf{k}) = J_0(\mathbf{k})$ for all $\mathbf{k} \in \mathcal{D}_J$, due to condition (i); all the better, $\tilde{J}_0$ is Schur-convex in $\mathcal{D}_J$ (a proof is given in Appendix D). Hence, for $\mathbf{k}$ and $\mathbf{q} \in \mathcal{D}_J$ such that $\mathbf{k}$ is majorized by $\mathbf{q}$, we have $J_0(\mathbf{k}) = \tilde{J}_0(\mathbf{k}) \leq \tilde{J}_0(\mathbf{q}) = J_0(\mathbf{q})$. The inequality (5.50) is guaranteed once $\mathbf{k}^*$ is majorized by any feasible $\mathbf{k}$, which is indeed true as shown below. For a feasible $\mathbf{k} = [k_1\ k_2 \cdots k_M]^T$ such that $k_1 \geq k_2 \geq \cdots \geq k_M$ without loss of generality. By Definition 5.4, it suffices to show

$$\sum_{m=1}^{s} k_m^* \leq \sum_{m=1}^{s} k_m \quad \text{for all} \quad 1 \leq s \leq M. \tag{5.54}$$

Assume otherwise that there exists $1 \leq s \leq M$ such that $\sum_{m=1}^{q} k_m^* \leq \sum_{m=1}^{q} k_m$, $1 \leq q \leq s-1$, whereas $\sum_{m=1}^{s} k_m^* > \sum_{m=1}^{s} k_m$. Then we have $k_s^* > k_s$, which together with $k_s^* \in \{q_t, q_t+1\}$ implies $q_t \geq k_m$, for $s+1 \leq m \leq M$, ending in the following contradiction:

$$\begin{aligned} k_t &= \sum_{m=1}^{s} k_m + \sum_{m=s+1}^{M} k_m \\ &\leq \sum_{m=1}^{s} k_m + \sum_{m=s+1}^{M} q_t \\ &\leq \sum_{m=1}^{s} k_m + \sum_{m=s+1}^{M} k_m^* \\ &< \sum_{m=1}^{s} k_m^* + \sum_{m=s+1}^{M} k_m^* = k_t. \end{aligned} \tag{5.55}$$

□

*C. Proof of Theorem 3.3*

Below we first derive an equivalent condition for the proposed stopping rule that is more amenable to analysis. Recall the residual $\mathbf{r}_m$ obtained in the $m$th iteration is the orthogonal projection of the previous residual $\mathbf{r}_{m-1}$ onto $\mathcal{R}\left(\mathbf{Y}_{\Lambda_{m-1}}\right)^{\perp}$. Since $\left\|\mathbf{r}_{m-1}^{(N)}\right\|_2^2 = \left\|\mathbf{r}_m^{(N)}\right\|_2^2 + \left\|\mathbf{r}_{m-1}^{(N)} - \mathbf{r}_m^{(N)}\right\|_2^2$, obtained from the Pythagorean theorem, the proposed stopping rule can be rewritten as

$$1 - \frac{\left\|\mathbf{r}_m^{(N)}\right\|_2}{\left\|\mathbf{r}_{m-1}^{(N)}\right\|_2} = 1 - \sqrt{1 - \left\|\tilde{\mathbf{r}}_m^{(N)}\right\|_2^2} \leq \sqrt{\frac{p}{n}}, \tag{5.56}$$

where

$$\tilde{\mathbf{r}}_m^{(N)} \triangleq \left(\mathbf{r}_{m-1}^{(N)} - \mathbf{r}_m^{(N)}\right) / \left\|\mathbf{r}_{m-1}^{(N)}\right\|_2 \in \mathcal{R}\left(\mathbf{Y}_{\Lambda_{m-1}}\right), \tag{5.57}$$

is the normalized difference of the residual vectors. Rearranging the inequality in (5.56) yields the following equivalent stopping condition

$$\left\|\tilde{\mathbf{r}}_m^{(N)}\right\|_2 \le \sqrt{2\sqrt{\frac{p}{n}}-\frac{p}{n}}\,. \tag{5.58}$$

Below we show that, with a high chance, (5.58) does not hold when $m \le \lceil d_L / p \rceil$ and is achieved (hence, GOMP stops) with $m = \lceil d_L / p \rceil + 1$. Formally, we prove the following inequalities

$$\left\|\tilde{\mathbf{r}}_m^{(N)}\right\|_2 > \sqrt{2\sqrt{p/n}-p/n}\,,\ \text{for all}\ \ 1 \le m \le \lceil d_L / p \rceil, \tag{5.59}$$

and

$$\left\|\tilde{\mathbf{r}}_{\lceil d_L/p\rceil+1}^{(N)}\right\|_2 \le \sqrt{2\sqrt{p/n}-p/n} \tag{5.60}$$

hold at once with a probability as high as claimed by Theorem 3.3.

To proceed, let $\mathbf{y}_j$ be a selected data vector in the $m$th iteration, $1 \le m \le \lceil d_L / p \rceil$. As long as (5.24), (5.28), (5.33), (5.34) and (5.40) hold, we have

$$\begin{aligned}
\left|\left\langle \mathbf{y}_j, \mathbf{r}_{m-1}^{(N)}\right\rangle\right| &\overset{(a)}{\ge} \left|\left\langle \mathbf{y}_{t_{k_m}}, \mathbf{r}_{m-1}^{(N)}\right\rangle\right| \overset{(b)}{\ge} \left|\left\langle \mathbf{x}_{\bar{t}_{k_m}}, \mathbf{r}_{m-1,\|}^{(N)}\right\rangle\right| - \max_{1\le j\le |\mathcal{Y}_L|-1-\left|\Lambda_{m-1}^t\right|}\left|\left\langle \mathbf{e}_{t_j}, \mathbf{r}_{m-1}^{(N)}\right\rangle\right| \\
&\overset{(c)}{\ge} \left[\frac{\left\|\mathbf{r}_{m-1,\|}^{(N)}\right\|_2}{\sqrt{d_L}} - \sqrt{\frac{6\log N}{n}}\left\|\mathbf{r}_{m-1}^{(N)}\right\|_2 \max_{1\le j\le |\mathcal{Y}_L|-1-\left|\Lambda_{m-1}^t\right|}\left\|\mathbf{e}_{t_j}\right\|_2\right] \\
&= \left\|\mathbf{r}_{m-1}^{(N)}\right\|_2 \left[\frac{\left\|\mathbf{r}_{m-1,\|}^{(N)}\right\|_2}{\sqrt{\left\|\mathbf{r}_{m-1,\perp}^{(N)}\right\|_2^2+\left\|\mathbf{r}_{m-1,\|}^{(N)}\right\|_2^2}}\frac{1}{\sqrt{d_L}} - \sqrt{\frac{6\log N}{n}} \max_{1\le j\le |\mathcal{Y}_L|-1-\left|\Lambda_{m-1}^t\right|}\left\|\mathbf{e}_{t_j}\right\|_2\right] \\
&\overset{(d)}{\ge} \left\|\mathbf{r}_{m-1}^{(N)}\right\|_2 \underbrace{\left[\frac{\sqrt{2}\,(1-3\sigma/2)}{\sqrt{\left(9d_L^2/4+4(1-3\sigma/2)^2 d_L\right)}} - \frac{3\sqrt{6\log N}\sigma}{2\sqrt{n}}\right]}_{\triangleq \eta},
\end{aligned} \tag{5.61}$$

in which (a) is true since $\mathbf{y}_j$ is selected as a neighbor, (b) follows form (5.9), (c) holds due to (5.24) and (5.28), and (d) is obtained by combining (5.33), (5.34), and (5.40). By definition (5.57) we have $\tilde{\mathbf{r}}_m^{(N)} = \mathrm{P}_{\mathcal{R}(\mathbf{Y}_{\Lambda_m})}\left(\mathbf{r}_{m-1}^{(N)} / \left\|\mathbf{r}_{m-1}^{(N)}\right\|_2\right)$, hence

$$\left\|\tilde{\mathbf{r}}_m^{(N)}\right\|_2 = \left\|\mathrm{P}_{\mathcal{R}(\mathbf{Y}_{\Lambda_m})}\left(\mathbf{r}_{m-1}^{(N)} / \left\|\mathbf{r}_{m-1}^{(N)}\right\|_2\right)\right\| \ge \left\|\mathrm{P}_{span\{\mathbf{y}_j\}}\left(\mathbf{r}_{m-1}^{(N)} / \left\|\mathbf{r}_{m-1}^{(N)}\right\|_2\right)\right\| = \left|\left\langle \mathbf{y}_j, \mathbf{r}_{m-1}^{(N)}\right\rangle\right| / \left(\left\|\mathbf{r}_{m-1}^{(N)}\right\|_2 \left\|\mathbf{y}_j\right\|_2\right), \tag{5.62}$$

which together with (5.61) implies

$$\left\|\tilde{\mathbf{r}}_m^{(N)}\right\|_2 \ge \frac{\left|\left\langle \mathbf{y}_j, \mathbf{r}_{m-1}^{(N)}\right\rangle\right|}{\left\|\mathbf{r}_{m-1}^{(N)}\right\|_2\left\|\mathbf{y}_j\right\|_2} \ge \frac{\eta\left\|\mathbf{r}_{m-1}^{(N)}\right\|_2}{\left\|\mathbf{r}_{m-1}^{(N)}\right\|_2\left\|\mathbf{y}_j\right\|_2} = \frac{\eta}{\left\|\mathbf{y}_j\right\|_2}. \tag{5.63}$$

With the aid of (5.63), inequality (5.59) is guaranteed once $\eta / \left\|\mathbf{y}_j\right\|_2 > \sqrt{2\sqrt{p/n}-p/n}$, which is typically true since the ambient dimension $n$ is drastically large.

As for (5.60), again by definition (5.57) we write

$$\tilde{\mathbf{r}}^{(N)}_{\lceil d_L/p\rceil+1} = \mathrm{P}_{\mathcal{R}\left(\mathbf{Y}_{\Lambda_{\lceil d_L/p\rceil+1}}\right)}(\mathbf{r}^{(N)}_{\lceil d_L/p\rceil} / \|\mathbf{r}^{(N)}_{\lceil d_L/p\rceil}\|_2). \tag{5.64}$$

Let $\{\mathbf{b}_1,\cdots,\mathbf{b}_{(\lceil d_L/p\rceil+1)p}\}$ be an orthonormal basis of $\mathcal{R}(\mathbf{Y}_{\Lambda_{\lceil d_L/p\rceil+1}})$, arranged in a way that $\{\mathbf{b}_{p+1},\cdots,\mathbf{b}_{(\lceil d_L/p\rceil+1)p}\}$ is an orthonormal basis of $\mathcal{R}(\mathbf{Y}_{\Lambda_{\lceil d_L/p\rceil}})$. Since $\mathbf{r}^{(N)}_{\lceil d_L/p\rceil}$ is orthogonal to $\mathcal{R}(\mathbf{Y}_{\Lambda_{\lceil d_L/p\rceil}})$, $\langle \mathbf{r}^{(N)}_{\lceil d_L/p\rceil}, \mathbf{b}_j\rangle = 0$ for all $p+1 \le j \le (\lceil d_L/p\rceil+1)p$. Thus, it follows

$$\begin{aligned}
\Pr\left\{\left\|\tilde{\mathbf{r}}^{(N)}_{\lceil d_L/q\rceil+1}\right\|_2 \le \sqrt{2\sqrt{\frac{p}{n}}-\frac{p}{n}}\right\} &= \Pr\left\{\left\|\mathrm{P}_{\mathcal{R}\left(\mathbf{Y}_{\Lambda_{\lceil d_L/p\rceil+1}}\right)}(\mathbf{r}^{(N)}_{\lceil d_L/p\rceil} / \|\mathbf{r}^{(N)}_{\lceil d_L/p\rceil}\|_2)\right\| \le \sqrt{2\sqrt{\frac{p}{n}}-\frac{p}{n}}\right\} \\
&\ge \Pr\left\{\bigcap_{k=1}^{p}\left\{\left\|\mathrm{P}_{span\{\mathbf{b}_k\}}(\mathbf{r}^{(N)}_{\lceil d_L/p\rceil} / \|\mathbf{r}^{(N)}_{\lceil d_L/p\rceil}\|_2)\right\| \le \sqrt{2\sqrt{\frac{p}{n}}-\frac{p}{n}}/\sqrt{p}\right\}\right\} \\
&= 1-\Pr\left\{\bigcup_{k=1}^{p}\left\{\left\|\mathrm{P}_{span\{\mathbf{b}_k\}}(\mathbf{r}^{(N)}_{\lceil d_L/p\rceil} / \|\mathbf{r}^{(N)}_{\lceil d_L/p\rceil}\|_2)\right\| > \sqrt{2\sqrt{\frac{p}{n}}-\frac{p}{n}}/\sqrt{p}\right\}\right\} \\
&\ge 1-\sum_{k=1}^{p}\Pr\left\{\left\|\mathrm{P}_{span\{\mathbf{b}_k\}}(\mathbf{r}^{(N)}_{\lceil d_L/p\rceil} / \|\mathbf{r}^{(N)}_{\lceil d_L/p\rceil}\|_2)\right\| > \sqrt{2\sqrt{\frac{p}{n}}-\frac{p}{n}}/\sqrt{p}\right\} \\
&= 1-\sum_{k=1}^{p}\Pr\left\{\left|\left\langle \frac{\mathbf{r}^{(N)}_{\lceil d_L/p\rceil}}{\|\mathbf{r}^{(N)}_{\lceil d_L/p\rceil}\|_2}, \mathbf{b}_k\right\rangle\right| > \sqrt{2\sqrt{\frac{p}{n}}-\frac{p}{n}}/\sqrt{p}\right\}..
\end{aligned} \tag{5.65}$$

According to our proof of Theorem 3.1, the neighbors selected in all $\lceil d_L/p\rceil$ iterations (i.e., all columns of $\mathbf{Y}_{\Lambda_{\lceil d_L/p\rceil}}$) are correct when (5.15), (5.18), (5.23), (5.24), (5.28), (5.33), (5.34) and (5.40) hold for all $1 \le m \le \lceil d_L/p\rceil$. If so, there then exists $\tilde{\mathbf{c}} \in \mathbb{R}^{p\lceil d_L/p\rceil}$ such that the "noiseless" signal vector $\mathbf{x}_N$ and neighbor matrix $\mathbf{X}_{\Lambda_{\lceil d_L/p\rceil}}$ satisfy

$$\mathbf{x}_N = \mathbf{X}_{\Lambda_{\lceil d_L/p\rceil}}\tilde{\mathbf{c}}. \tag{5.66}$$

Noting that $\mathbf{Y}_{\Lambda_{\lceil d_L/p\rceil}} = \mathbf{X}_{\Lambda_{\lceil d_L/p\rceil}} + \mathbf{E}_{\Lambda_{\lceil d_L/p\rceil}}$, where $\mathbf{E}_{\Lambda_{\lceil d_L/p\rceil}}$ is the noise matrix, the residual vector at the $\lceil d_L/p\rceil$th iteration can be written as

$$\begin{aligned}
\mathbf{r}^{(N)}_{\lceil d_L/p\rceil} &= \left(\mathbf{I}-\mathbf{Y}_{\Lambda_{\lceil d_L/p\rceil}}\mathbf{Y}^{\dagger}_{\Lambda_{\lceil d_L/p\rceil}}\right)\mathbf{y}_N \\
&= (\mathbf{I}-\mathbf{Y}_{\Lambda_{\lceil d_L/p\rceil}}\mathbf{Y}^{\dagger}_{\Lambda_{\lceil d_L/p\rceil}})\mathbf{x}_N + \left(\mathbf{I}-\mathbf{Y}_{\Lambda_{\lceil d_L/p\rceil}}\mathbf{Y}^{\dagger}_{\Lambda_{\lceil d_L/p\rceil}}\right)\mathbf{e}_N \\
&= (\mathbf{I}-\mathbf{Y}_{\Lambda_{\lceil d_L/p\rceil}}\mathbf{Y}^{\dagger}_{\Lambda_{\lceil d_L/p\rceil}})\left(\mathbf{x}_N-\mathbf{Y}_{\Lambda_{\lceil d_L/p\rceil}}\tilde{\mathbf{c}}\right) + \left(\mathbf{I}-\mathbf{Y}_{\Lambda_{\lceil d_L/p\rceil}}\mathbf{Y}^{\dagger}_{\Lambda_{\lceil d_L/p\rceil}}\right)\mathbf{e}_N \\
&= (\mathbf{I}-\mathbf{Y}_{\Lambda_{\lceil d_L/p\rceil}}\mathbf{Y}^{\dagger}_{\Lambda_{\lceil d_L/p\rceil}})\mathbf{E}_{\Lambda_{\lceil d_L/p\rceil}}\tilde{\mathbf{c}} + \left(\mathbf{I}-\mathbf{Y}_{\Lambda_{\lceil d_L/p\rceil}}\mathbf{Y}^{\dagger}_{\Lambda_{\lceil d_L/p\rceil}}\right)\mathbf{e}_N \\
&= (\mathbf{I}-\mathbf{Y}_{\Lambda_{\lceil d_L/p\rceil}}\mathbf{Y}^{\dagger}_{\Lambda_{\lceil d_L/p\rceil}})\left(\mathbf{E}_{\Lambda_{\lceil d_L/p\rceil}}\tilde{\mathbf{c}}+\mathbf{e}_N\right),
\end{aligned} \tag{5.67}$$

which is the projection of the Gaussian random vector $\mathbf{E}_{\Lambda_{\lceil d_L/p\rceil}}\tilde{\mathbf{c}}+\mathbf{e}_N$ onto $\mathcal{R}\left(\mathbf{Y}_{\Lambda_{\lceil d_L/p\rceil}}\right)^{\perp}$. Being a linear combination of rotationally invariant vectors, $\mathbf{E}_{\Lambda_{\lceil d_L/p\rceil}}\tilde{\mathbf{c}}+\mathbf{e}_N$ remains so: this implies the projection $\mathbf{r}^{(N)}_{\lceil d_L/p\rceil}$ is also rotationally invariant on $\mathcal{R}\left(\mathbf{Y}_{\Lambda_{\lceil d_L/p\rceil}}\right)^{\perp}$. Then using part (a) of [10, Ex. 5.25] with $\mathbf{a} = \mathbf{r}^{(N)}_{\lceil d_L/p\rceil} / \|\mathbf{r}^{(N)}_{\lceil d_L/p\rceil}\|_2$ and $\mathbf{b} = \mathbf{b}_k$, $1 \le k \le p$, we can obtain

$$\Pr\left\{\left|\left\langle \frac{\mathbf{r}^{(N)}_{\lceil d_L/p\rceil}}{\left\|\mathbf{r}^{(N)}_{\lceil d_L/p\rceil}\right\|_2}, \mathbf{b}_k \right\rangle\right| > \sqrt{2\sqrt{\frac{p}{n}} - \frac{p}{n}} / \sqrt{p}\right\} \le 2e^{-\left(\sqrt{2\sqrt{\frac{p}{n}}-\frac{p}{n}}/\sqrt{p}\right)^2 \frac{n-p(\lceil d_L/p\rceil)}{2}} \approx 2e^{-\sqrt{n/p}}, \tag{5.68}$$

which together with (5.65) implies (5.60) holds with a probability at least $1-2pe^{-\sqrt{n/p}}$. In summary, once the inequalities (5.15), (5.18), (5.23), (5.24), (5.28), (5.33), (5.34) and (5.40) hold for all $1 \le m \le \lceil d_L / p \rceil$, then (5.59) and (5.60) hold with a probability at least $1-2pe^{-\sqrt{n/p}}$. The proof is then completed by employing the union bound. □

# VI. CONCLUSION

Fast acquisition of many true neighbors underlies the success of SSC in real-world applications. Under the framework of greedy selection, which is pretty suited for low-complexity implementation, we propose a GOMP sparse regression scheme, together with a novel subspace-dimensionality-aware stopping rule, to boost neighbor recovery. Thanks to multiple neighbor identification per iteration, the proposed GOMP involves fewer iterations, thereby enjoying even lower algorithmic complexity than conventional OMP; in addition, the residual vector better stands up to noise corruption, consequently bringing about higher neighbor identification accuracy. Our proposed stopping criterion is appealing in that it depends entirely on a knowledge of the ambient space dimension, in marked contrast with the existing solution [20] that requires extra off-line estimation of either subspace dimension or noise power. Besides algorithm development, in-depth neighbor recovery rate analyses were conducted to justify the merits of the proposed GOMP; the obtained analytic results are further validated by computations using both synthetic and real human face data sets. Overall, our study presents a new greedy-based SSC scheme, with provable performance guarantees, that can pave the way for practical applications. Future work will analyze AoD of both GOMP and OMP under the considered semi-random model, particularly to show GOMP enjoys a smaller average AoD; an analytic study of AoD would further offer certain guidelines on determining the optimum number of neighbors to recover in each iteration, for the purpose of improving the current scheme, which identifies a constant number $(p \ge 1)$ of neighbors throughout all iterations, in a more dynamic environment.

# APPENDIX

## A. Proof of Lemma 5.1

Lemma 5.1 is a straightforward extension of the following results.

***Lemma A.1 ([10, Ex. 5.25] and [18, eq. (59)]):*** Let $\mathbf{a} \in \mathbb{R}^m$ be a random vector uniformly distributed over the unit-sphere of $\mathbb{R}^m$. Then for $\varepsilon \geq 0$ and a fixed $\mathbf{b} \in \mathbb{R}^m$, we have

(a) $\Pr\left\{\left|\mathbf{a}^T\mathbf{b}\right| > \varepsilon \left\|\mathbf{b}\right\|_2\right\} \leq 2e^{-m\varepsilon^2/2}$, (A.1)

(b) $\Pr\left\{\left|\mathbf{a}^T\mathbf{b}\right| < \varepsilon \left\|\mathbf{b}\right\|_2 / \sqrt{m}\right\} \leq \sqrt{2/\pi}\varepsilon$. (A.2)

□

We commence with the proof of part (a). Let $f_{\mathbf{b}} : \mathbb{R}^m \to \mathbb{R}$ be the probability density function of random vector $\mathbf{b}$. Using (A.1) we have

$$\begin{aligned}\Pr\left\{\left|\mathbf{a}^T\mathbf{b}\right| > \varepsilon \left\|\mathbf{b}\right\|_2\right\} &= \int_{\mathbb{R}^m} \Pr\left\{\left|\mathbf{a}^T\mathbf{b}\right| > \varepsilon \left\|\mathbf{b}\right\|_2 \middle| \mathbf{b} = \mathbf{v}\right\} f_{\mathbf{b}}(\mathbf{v})\, d\mathbf{v} \\ &\leq \int_{\mathbb{R}^m} 2e^{-m\varepsilon^2/2} f_{\mathbf{b}}(\mathbf{v})\, d\mathbf{v} \\ &= 2e^{-m\varepsilon^2/2} \int_{\mathbb{R}^m} f_{\mathbf{b}}(\mathbf{v})\, d\mathbf{v} \\ &= 2e^{-m\varepsilon^2/2}.\end{aligned} \tag{A.3}$$

Similarly, for part (b) it follows

$$\begin{aligned}\Pr\left\{\left|\mathbf{a}^T\mathbf{b}\right| < \varepsilon \left\|\mathbf{b}\right\|_2 / \sqrt{m}\right\} &= \int_{\mathbb{R}^m} \Pr\left\{\left|\mathbf{a}^T\mathbf{b}\right| < \varepsilon \left\|\mathbf{b}\right\|_2 / \sqrt{m} \middle| \mathbf{b} = \mathbf{v}\right\} f_{\mathbf{b}}(\mathbf{v})\, d\mathbf{v} \\ &\overset{(a)}{\leq} \int_{\mathbb{R}^m} \sqrt{2/\pi}\varepsilon f_{\mathbf{b}}(\mathbf{v})\, d\mathbf{v} \\ &= \sqrt{2/\pi}\varepsilon \int_{\mathbb{R}^m} f_{\mathbf{b}}(\mathbf{v})\, d\mathbf{v} \\ &= \sqrt{2/\pi}\varepsilon,\end{aligned} \tag{A.4}$$

where (a) follows from (A.2). The proof of Lemma 5.1 is completed. □

## B. Proof of Lemma 5.2

The following lemma is needed for deriving Lemma 5.2.

***Lemma B.1 [12, p-2231]:*** Let $\mathbf{x} = [x_1\ x_2 \cdots x_m]^T \in \mathbb{R}^m$ be a random vector uniformly distributed over the unit sphere of $\mathbb{R}^m$ and $\mathbf{A} = [a_{ij}] \in \mathbb{R}^{n\times m}$ be given. Then the expectation $E_u\left[\left\|\mathbf{A}\mathbf{x}\right\|_2^2\right]$, where the subscript $u$ stands for the uniform distribution, is given by

$$E_u\left[\left\|\mathbf{A}\mathbf{x}\right\|_2^2\right] = \left\|\mathbf{A}\right\|_F^2 / m. \tag{B.1}$$

□

Assume $\mathbf{x}_i \in \mathcal{S}_l$, $l \neq L$. Then we have

$$\begin{aligned}
&\Pr\left(\left\{\left\|\mathbf{U}_L^T\mathbf{x}_i\right\|_2 > 4\log N \frac{\left\|\mathbf{U}_L^T\mathbf{U}_l\right\|_F}{\sqrt{d_L d_l}}\right\}\right) \\
&\overset{(a)}{=} \Pr\left(\left\{\left\|\mathbf{U}_L^T\mathbf{x}_i\right\|_2 > \frac{4\log N}{\sqrt{d_L}}\sqrt{E_u\left[\left\|\mathbf{U}_L^T\mathbf{U}_l\mathbf{a}\right\|_2^2\right]}\right\}\right) \\
&\overset{(b)}{\leq} \frac{2}{N^{(8\log N)/d_L}},
\end{aligned} \tag{B.2}$$

where (a) is true by introducing an “auxiliary” random vector $\mathbf{a}$ uniformly distributed over the unit sphere of $\mathbb{R}^{d_l}$ and using Lemma B.1, and (b) holds by the same procedures for the proof of [12, Lemma 7.5]. Since $\mathbf{r}_{m,\|}^{(N)} \in \mathcal{S}_L$, we have $\mathbf{r}_{m,\|}^{(N)} / \left\|\mathbf{r}_{m,\|}^{(N)}\right\|_2 = \mathbf{U}_L\mathbf{z}$ with $\|\mathbf{z}\|_2 = 1$. The Cauchy-Schwartz inequality implies

$$\left|\left\langle \frac{\mathbf{r}_{m,\|}^{(N)}}{\left\|\mathbf{r}_{m,\|}^{(N)}\right\|_2}, \mathbf{x}_i \right\rangle\right| = \left|\mathbf{z}^T\mathbf{U}_L^T\mathbf{x}_i\right| \leq \|\mathbf{z}\|_2 \left\|\mathbf{U}_L^T\mathbf{x}_i\right\|_2 = \left\|\mathbf{U}_L^T\mathbf{x}_i\right\|_2, \tag{B.3}$$

and therefore

$$\left\{\left|\left\langle \mathbf{x}_i, \frac{\mathbf{r}_{m,\|}^{(N)}}{\left\|\mathbf{r}_{m,\|}^{(N)}\right\|_2} \right\rangle\right| > 4\log N \frac{\left\|\mathbf{U}_L^T\mathbf{U}_l\right\|_F}{\sqrt{d_L d_l}}\right\} \subset \left\{\left\|\mathbf{U}_L^T\mathbf{x}_i\right\|_2 > 4\log N \frac{\left\|\mathbf{U}_L^T\mathbf{U}_l\right\|_F}{\sqrt{d_L d_l}}\right\}. \tag{B.4}$$

With (B.2) and (B.4), it follows immediately

$$\Pr\left\{\left|\left\langle \mathbf{x}_i, \frac{\mathbf{r}_{m,\|}^{(N)}}{\left\|\mathbf{r}_{m,\|}^{(N)}\right\|_2} \right\rangle\right| > 4\log N \frac{\left\|\mathbf{U}_L^T\mathbf{U}_l\right\|_F}{\sqrt{d_L d_l}}\right\} \leq \frac{2}{N^{(8\log N)/d_L}}. \tag{B.5}$$

The proof of Lemma 5.2 is completed. □

*C. Proof of Lemma 5.3*

Let

$$\mathcal{D}(\mathbf{v}) \triangleq \left\{\left(\mathbf{y}_1, \mathbf{y}_2, \cdots, \mathbf{y}_N\right) : \frac{\mathbf{r}_{m-1,\perp}^{(N)}}{\left\|\mathbf{r}_{m-1,\perp}^{(N)}\right\|_2} = \mathbf{v}\right\}, \tag{C.1}$$

$f : \mathbb{R}^n \times \cdots \times \mathbb{R}^n \to \mathbb{R}$ be the probability density function of $\left(\mathbf{y}_1, \mathbf{y}_2, \cdots, \mathbf{y}_N\right)$, and $\mathbb{S}^{n-1}$ be the unit-sphere of $\mathbb{R}^n$. Then for $\mathbf{z} \in \mathcal{S}_L^\perp \cap \mathbb{S}^{n-1}$, we have

$$\Pr\left\{\left|\left\langle \mathbf{r}_{m-1,\perp}^{(N)} / \left\|\mathbf{r}_{m-1,\perp}^{(N)}\right\|_2, \mathbf{z}\right\rangle\right| > \varepsilon\right\}$$
$$= \int_{\left(\mathcal{S}_L^{\perp}\cap\mathbb{S}^{n-1}\right)\cap\left\{\mathbf{r}:\left|\langle\mathbf{r},\mathbf{z}\rangle\right|>\varepsilon\right\}} \left[\int_{\mathcal{D}(\mathbf{v})} f(\mathbf{y}_1,\mathbf{y}_2,\cdots,\mathbf{y}_N)d\mathbf{y}_1 d\mathbf{y}_2\cdots d\mathbf{y}_N\right] d\mathbf{v} \tag{C.2}$$
$$= \int_{\left(\mathcal{S}_L^{\perp}\cap\mathbb{S}^{n-1}\right)\cap\left\{\mathbf{r}:\left|\langle\mathbf{r},\mathbf{z}\rangle\right|>\varepsilon\right\}} \left[\int_{\mathcal{K}\cap\mathcal{D}(\mathbf{v})} f(\mathbf{y}_1,\mathbf{y}_2,\cdots,\mathbf{y}_N)d\mathbf{y}_1 d\mathbf{y}_2\cdots d\mathbf{y}_N + \int_{\mathcal{K}^c\cap\mathcal{D}(\mathbf{v})} f(\mathbf{y}_1,\mathbf{y}_2,\cdots,\mathbf{y}_N)d\mathbf{y}_1 d\mathbf{y}_2\cdots d\mathbf{y}_N\right] d\mathbf{v}$$

where the last equality holds by defining

$$\mathcal{K} \triangleq \left\{\mathbf{x}+\mathbf{e}:\mathbf{x}\in\mathcal{S}_{k_1}\cap\mathbb{S}^{n-1},\ \|\mathbf{e}\|_2\le 3\sigma/2\right\}\times\left\{\mathbf{x}+\mathbf{e}:\mathbf{x}\in\mathcal{S}_{k_2}\cap\mathbb{S}^{n-1},\ \|\mathbf{e}\|_2\le 3\sigma/2\right\}\times$$
$$\cdots\times\left\{\mathbf{x}+\mathbf{e}:\mathbf{x}\in\mathcal{S}_{k_{N-1}}\cap\mathbb{S}^{n-1},\ \|\mathbf{e}\|_2\le 3\sigma/2\right\}\times\left\{\mathbf{x}+\mathbf{e}:\mathbf{x}\in\mathcal{S}_{L}\cap\mathbb{S}^{n-1},\ \|\mathbf{e}\|_2\le 3\sigma/2\right\} \tag{C.3}$$

in which we assume that $\mathbf{y}_i \in \mathcal{Y}_{k_i}$, hence $\mathbf{x}_i \in \mathcal{S}_{k_i}$, for $1\le i\le N-1$. Since the maximum function is continuous and composition of continuous functions is also continuous, $\mathbf{r}_{m-1,\perp}^{(N)} / \left\|\mathbf{r}_{m-1,\perp}^{(N)}\right\|_2 = \mathrm{P}_{\mathcal{S}_L^{\perp}}\left(\mathbf{I}-\mathbf{Y}_{\Lambda_{m-1}}\mathbf{Y}_{\Lambda_{m-1}}^{\dagger}\right)\mathbf{y}_N / \left\|\mathrm{P}_{\mathcal{S}_L^{\perp}}\left(\mathbf{I}-\mathbf{Y}_{\Lambda_{m-1}}\mathbf{Y}_{\Lambda_{m-1}}^{\dagger}\right)\mathbf{y}_N\right\|_2$ is a continuous function of $(\mathbf{y}_1,\mathbf{y}_2,\cdots,\mathbf{y}_N)$. Notice that $\mathcal{K}$ is closed and bounded in the Euclidean space, and is therefore compact by the Heine-Borel theorem [29]. Since $f(\mathbf{y}_1,\mathbf{y}_2,\cdots,\mathbf{y}_N)$ is continuous on the compact set $\mathcal{K}$, by the extreme value theorem [29] there exists a constant $\delta$ such that

$$f(\mathbf{y}_1,\mathbf{y}_2,\cdots,\mathbf{y}_N)\le\delta,\ \text{for all}\ \left(\mathbf{y}_1,\mathbf{y}_2,\cdots\mathbf{y}_N\right)\in\mathcal{K}, \tag{C.4}$$

which together with (C.2) implies

$$\Pr\left\{\left|\left\langle \mathbf{r}_{m-1,\perp}^{(N)} / \left\|\mathbf{r}_{m-1,\perp}^{(N)}\right\|_2, \mathbf{z}\right\rangle\right| > \varepsilon\right\}$$
$$\le \int_{\left(\mathcal{S}_L^{\perp}\cap\mathbb{S}^{n-1}\right)\cap\left\{\mathbf{r}:\left|\langle\mathbf{r},\mathbf{z}\rangle\right|>\varepsilon\right\}} \left[\int_{\mathcal{K}\cap\mathcal{D}(\mathbf{v})} \delta d\mathbf{y}_1 d\mathbf{y}_2\cdots d\mathbf{y}_N + \int_{\mathcal{K}^c\cap\mathcal{D}(\mathbf{v})} f(\mathbf{y}_1,\mathbf{y}_2,\cdots,\mathbf{y}_N)d\mathbf{y}_1 d\mathbf{y}_2\cdots d\mathbf{y}_N\right] d\mathbf{v}$$
$$\le \int_{\left(\mathcal{S}_L^{\perp}\cap\mathbb{S}^{n-1}\right)\cap\left\{\mathbf{r}:\left|\langle\mathbf{r},\mathbf{z}\rangle\right|>\varepsilon\right\}} \left[\int_{\mathcal{K}} \delta d\mathbf{y}_1 d\mathbf{y}_2\cdots d\mathbf{y}_N + \int_{\mathcal{K}^c\cap\mathcal{D}(\mathbf{v})} f(\mathbf{y}_1,\mathbf{y}_2,\cdots,\mathbf{y}_N)d\mathbf{y}_1 d\mathbf{y}_2\cdots d\mathbf{y}_N\right] d\mathbf{v}$$
$$\overset{(a)}{\le} \int_{\left(\mathcal{S}_L^{\perp}\cap\mathbb{S}^{n-1}\right)\cap\left\{\mathbf{r}:\left|\langle\mathbf{r},\mathbf{z}\rangle\right|>\varepsilon\right\}} \left[\delta\left|\mathcal{K}\right| + Ne^{-n/8}\right] d\mathbf{v}$$
$$\overset{(b)}{=} \int_{\left(\mathcal{S}_L^{\perp}\cap\mathbb{S}^{n-1}\right)\cap\left\{\mathbf{r}:\left|\langle\mathbf{r},\mathbf{z}\rangle\right|>\varepsilon\right\}} \left[c / A(\mathcal{S}_L^{\perp}\cap\mathbb{S}^{n-1})\right] d\mathbf{v}$$
$$= c \int_{\left(\mathcal{S}_L^{\perp}\cap\mathbb{S}^{n-1}\right)\cap\left\{\mathbf{r}:\left|\langle\mathbf{r},\mathbf{z}\rangle\right|>\varepsilon\right\}} 1/ A(\mathcal{S}_L^{\perp}\cap\mathbb{S}^{n-1}) d\mathbf{v} \tag{C.5}$$
$$\overset{(c)}{\le} 2ce^{-(n-d_L)\varepsilon^2/2},$$

where (a) holds by [20, Lemma 10] with $|\mathcal{K}|$ here denoting the Lebesgue measure of $\mathcal{K}$, (b) is true as we define $A(\mathcal{S}_L^{\perp} \cap \mathbb{S}^{n-1})$ to be the surface area of $\mathcal{S}_L^{\perp} \cap \mathbb{S}^{n-1}$ and the constant $c \triangleq \left(\delta|\mathcal{K}| + Ne^{-n/8}\right)A(\mathcal{S}_L^{\perp} \cap \mathbb{S}^{n-1})$, and (c) follows from [10, Ex. 5.25]. With (C.5), we reach

$$\begin{aligned}
\Pr\left\{\left|\left\langle \mathbf{r}_{m-1,\perp}^{(N)} / \left\|\mathbf{r}_{m-1,\perp}^{(N)}\right\|_2, \mathbf{x}_i \right\rangle\right| > \varepsilon\right\} &= \int_{\mathbb{R}^n} \Pr\left\{\left|\left\langle \mathbf{r}_{m-1,\perp}^{(N)} / \left\|\mathbf{r}_{m-1,\perp}^{(N)}\right\|_2, \mathbf{x}_i \right\rangle\right| > \varepsilon \,\middle|\, \mathbf{x}_i = \mathbf{v}\right\} f_{\mathbf{x}_i}(\mathbf{v})\, d\mathbf{v} \\
&\overset{(a)}{\leq} \int_{\mathbb{R}^n} 2ce^{-(n-d_L)\varepsilon^2/2} f_{\mathbf{x}_i}(\mathbf{v})\, d\mathbf{v} \\
&= 2ce^{-(n-d_L)\varepsilon^2/2},
\end{aligned} \tag{C.6}$$

where (a) holds by (C.5), in which $f_{\mathbf{x}_i}$ is the probability density function of $\mathbf{x}_i$. Let $\varepsilon = \sqrt{6\log N/(n-d_L)}$. Then (C.6) gives

$$\Pr\left\{\left|\left\langle \mathbf{r}_{m-1,\perp}^{(N)} / \left\|\mathbf{r}_{m-1,\perp}^{(N)}\right\|_2, \mathbf{x}_i \right\rangle\right| > \sqrt{6\log N/(n-d_L)}\right\} \leq \frac{2c}{N^3}, \tag{C.7}$$

which completes the proof. □

*D. Proof of Schur-Convexity of* $\tilde{J}_0$

Notably, straightforward manipulations show

$$(k_m - k_q)\left(\frac{\partial J_0}{\partial k_m} - \frac{\partial J_0}{\partial k_q}\right) = (k_m - k_q)\left(J'(k_m) - J'(k_q)\right) \geq 0 \quad \text{for all} \quad k_m, k_q \in \{1, 2, \cdots, p\}. \tag{D.1}$$

It then follows

$$(k_m - k_q)\left(\frac{\partial \tilde{J}_0}{\partial k_m} - \frac{\partial \tilde{J}_0}{\partial k_q}\right) = (k_m - k_q)\left(\tilde{J}'(k_m) - \tilde{J}'(k_q)\right) \overset{(a)}{=} (k_m - k_q)\left(J'(k_m) - J'(k_q)\right) \overset{(b)}{\geq} 0$$

$$\text{for all} \quad k_m, k_q \in \{1, 2, \cdots, p\}, \tag{D.2}$$

where (a) holds by (5.51) and (b) is due to (D.1). When evaluated around $k_m \in \{1, 2, \cdots, p\}$ and $k_q = 0$, the resultant partial derivative reads

$$\begin{aligned}
(k_m - k_q)\left(\frac{\partial \tilde{J}_0}{\partial k_m} - \frac{\partial \tilde{J}_0}{\partial k_q}\right) &= k_m\left(\tilde{J}'(k_m) - \tilde{J}'(0)\right) \\
&\overset{(a)}{=} k_m\left(J'(k_m) - \tilde{J}'(0)\right) \\
&\overset{(b)}{=} k_m\left(J'(k_m) - J'(1)\right) \\
&\overset{(c)}{\geq} 0,
\end{aligned} \tag{D.3}$$

where (a) holds by (5.51), (b) follows from (5.52), and (c) is true due to (D.1). Clearly, the function $\tilde{J}_0$ is permutation invariant, together with (D.2) and (D.3) ensures that $\tilde{J}_0$ satisfies the conditions of Lemma 5.5, and thus is Schur-convex. □